%% file: colm2024_conference.tex
\pdfoutput=1

\documentclass[11pt]{article}

\usepackage{colm2024_conference}
\colmfinaltrue

\usepackage{pgfplots, pgfplotstable}
\usepackage{amssymb}
\usepackage{mathtools}
\usepackage[thinc]{esdiff}
\usepackage{xspace}
\usepackage{tensor}
\usepackage{amsthm}
\usepackage{multirow}
\usepackage{bm}
\usepackage{thmtools}
\usepackage{thm-restate}
\usepackage[noabbrev,capitalize,nameinlink]{cleveref} %

\crefname{equation}{equation}{equations}   %
\crefname{footnote}{footnote}{footnotes}   %
\crefname{section}{\S}{\S\S}
\Crefname{section}{\S}{\S\S}    %
\crefformat{section}{#2\S#1#3}  %
\Crefformat{section}{#2\S#1#3}
\crefrangeformat{section}{\S\S#3#1#4--#5#2#6}
\Crefrangeformat{section}{\S\S#3#1#4--#5#2#6}
\crefformat{section}{#2\S#1#3}  %
\crefrangeformat{section}{\S\S#3#1#4--#5#2#6}

\usepackage{tikz}
\usetikzlibrary{automata, positioning, arrows}
\usepackage{pgfplots}
\pgfplotsset{compat=1.14}
\usepackage{floatrow}

\usepackage{newtxtext}
\usepackage{newtxmath}

\usepackage[activate=true,
final,
tracking=false,   %
kerning=true,
spacing=true,
factor=1050,
stretch=30,
shrink=30]{microtype}      %

\usepackage{todonotes}
\makeatletter
\newcommand*\iftodonotes{\if@todonotes@disabled\expandafter\@secondoftwo\else\expandafter\@firstoftwo\fi}  %
\makeatother

\newcommand{\mycomment}[1]{%
}%

\newcommand{\cutforspace}[1]{%
}%

\usepackage{bm}

\newcommand{\vecc}[1]{{\boldsymbol{\mathbf{#1}}}}

\usepackage{stmaryrd}  %

\usepackage[safe]{tipa}

\newcommand{\Real}{\mathbb{R}}

\usepackage{url}

\newcommand{\approptoinn}[2]{\mathrel{\vcenter{
			\offinterlineskip\halign{\hfil$##$\cr
				#1\propto\cr\noalign{\kern2pt}#1\sim\cr\noalign{\kern-2pt}}}}}

\crefname{ineq}{inequality}{inequalities}
\creflabelformat{ineq}{#2{\upshape(#1)}#3}

\numberwithin{locallemma}{subsection}

\usepackage{siunitx}
\usepackage{comment}

\usepackage{enumitem}

\usepackage[T1]{fontenc}    %
\usepackage{url}            %
\usepackage{booktabs}       %
\usepackage{amsfonts}       %
\usepackage{nicefrac}       %

\usepackage{multirow}
\usepackage{arydshln}
\usepackage{graphicx}
\usepackage{amssymb}  %
\usepackage{subcaption}
\usepackage{amsmath} %

\usepackage{diagbox}
\usepackage{subcaption}

\usepackage{color, colortbl}

\usepackage{pifont}%

\usepackage{url}

\usepackage{tabularx}
\usepackage{colortbl}
\usepackage{tikz} 
\usepackage{adjustbox}

\usepackage{makecell}

\usepackage{xspace}
\newcommand\model{FLix\xspace}

\usepackage[ruled,vlined]{algorithm2e}

\SetCommentSty{commfont}

\DeclareRobustCommand{\thinskip}{\hskip 0.16667em\relax}
\def\emdash{---}
\def\d@sh#1#2{\unskip#1\thinskip#2\thinskip\ignorespaces}
\def\Dash{\d@sh\nobreak\emdash}
\def\Ldash{\d@sh\empty{\hbox{\emdash}\nobreak}}
\def\Rdash{\d@sh\nobreak\emdash}

\title{Inducing Generalization across Languages and Tasks \\ using Featurized Low-Rank Mixtures}

\author{Chu-Cheng Lin\thanks{equal contribution, \texttt{\{kitsing,xinyiwang\}@google.com}} , Xinyi Wang\footnotemark[1] , Jonathan H. Clark,  Han Lu, \And
Yun Zhu, Chenxi Whitehouse, Hongkun Yu \\ \\
Google}

\begin{document}
\maketitle

\begin{abstract}
Adapting pretrained large language models~(LLMs) to various downstream tasks in tens or hundreds of human languages is computationally expensive. 
Parameter-efficient fine-tuning~(PEFT) significantly reduces the adaptation cost, by tuning only a small amount of parameters. However, common PEFT methods LoRA~\citep{hu2022lora} suffer from suboptimal performance on diverse dataset mixtures, due to aggressive parameter tying and negative interference among different datasets. 
In this work, we propose {\bf F}eaturized {\bf L}ow-rank  M{\bf ix}tures~(\model), a novel PEFT method designed for effective multitask multilingual adaptation. 
\model associates each unique dataset feature, such as the dataset's language or task, with its own low-rank weight update parameters. 
By composing feature-specific parameters for each dataset, \model can accommodate diverse dataset mixtures and generalize better to unseen datasets. Our experiments show that \model leads to significant improvements over a variety of tasks for both supervised learning and zero-shot settings with gains of up to $14.2$ in exact match points in zero-shot semantic parsing.
\end{abstract}

\input{intro_v2}

\input{background_v2}

\input{method_v2}

\input{experiment}

\input{ablations}

\input{related_work}

\input{conclusion}

\bibliographystyle{colm2024_conference}
\bibliography{colm2024_conference,custom}

\clearpage
\appendix
\input{appendix}

\end{document}

%% file: intro_v2.tex
\section{Introduction}

\label{sec:intro}

Large language models~(LLMs) have shown impressive performance on various real world applications in many different human languages \citep{NEURIPS2020_1457c0d6,Soltan2022AlexaTM2F,anil2023palm}.
While there have been notable successes aligning an LLM to become a \emph{generalist} which can follow human instructions to perform different tasks \citep{Ouyang2022TrainingLM}, there are also significant interests in adapting an LLM into \emph{specialists}, each of which works on its specific task that is known \emph{a priori}.

Intuitively, LLM adaptation can be done by continued training (or \emph{fine-tuning}) on target languages and datasets. However, fine-tuning all model parameters on every dataset can be computationally and financially prohibitive. 
Parameter-efficient fine-tuning methods~(PEFT) such as LoRA \citep{hu2022lora} and prompt tuning \citep{lester-etal-2021-power} reduce the computational costs of adapting LLMs to a downstream task. They parametrize LLM fine-tuning with a small set of trainable parameters, keeping the majority of LLM parameters frozen.

While PEFT has been widely used to adapt LLMs to a single dataset, very few prior works studied the best practices for adapting the model jointly on many different use cases.
\cite{vu2022overcoming} proposed adding a multilingual pretraining stage to prompt tuning using multilingual unlabeled data to improve zero-shot summarization. \cite{wang2023multitask} proposed a multitask prompt tuning method that learns a single soft prompt which could be used to be adapt to other target tasks. However, these methods only considered either multilingual or multitask datasets, while we generally would like to adapt the model such that it generalizes along multiple axes~(i.e. tasks and languages). Moreover, they require multiple tuning stages, which limits their applicability in practice.

In this paper, we propose {\bf F}eaturized {\bf L}ow-rank M{\bf ix}tures (\model), an extension of LoRA for modeling diverse dataset mixtures. Compared to LoRA which applies the same low-rank adaptation for all inputs, \model parametrizes such updates to \emph{decompose} linearly as a sum of feature-specific low-rank adaptations, each associated with an active dataset feature, such as language or task ID.
Under \model, different adaptations can be learned for different features. The compositional nature of \model also provides an inductive bias for learning generalizable adaptations. Moreover,
\model is generally computationally efficient: it only needs to activate a tiny fraction of trainable parameters for each input, making both tuning and deployment efficient.
\model is related to prior works on sub-network composition~\citep{lin-etal-2021-learning,ilharco2022editing}, which show that models fine-tuned on different tasks could be composed together as a single model. 

In this article, we contribute:
\begin{itemize}
    \item a modeling formulation that disentangles learning of tasks and languages in a way that improves generalization quality while remaining highly efficient;
    \item experimental evidence that the model improves generalization quality across four very different tasks: named entity recognition, semantic parsing, in-language question answering, and cross-lingual question answering; and
    \item evidence for the hypothesis that imbuing models with meta-data such as task and language improves quality, even when compared with powerful modern adaptation methods such as LoRA.
\end{itemize}

The rest of this paper is structured as follows: In \cref{sec:background}, we first formulate the multitask multilingual learning setting, and discuss the challenges associated with the current PEFT methods. 
In \cref{sec:modularized-multitask-multilingual-lora}, we describe the design of \model, and how \model adapts to zero-shot scenarios.
We also propose to train \model with feature dropout, which encourages positive transfer and generalization. We evaluate \model on multitask or multilingual learning setting~(\cref{sec:mt-or-ml}), and later on joint multilingual multitask tuning and zero-shot generalization~(\cref{sec:joint-mtml}). The experiment results and ablations show that \model brings significant improvements over standard PEFT methods for all settings, and it is especially effective when used with very diverse training mixture and at zero-shot generalizations (\cref{sec:experiment-results}, \cref{sec:ablations}).

%% file: background_v2.tex
\begin{figure*}
    \centering
    \begin{subfigure}{0.45\textwidth}
    \centering
    \includegraphics[width=.8\textwidth]{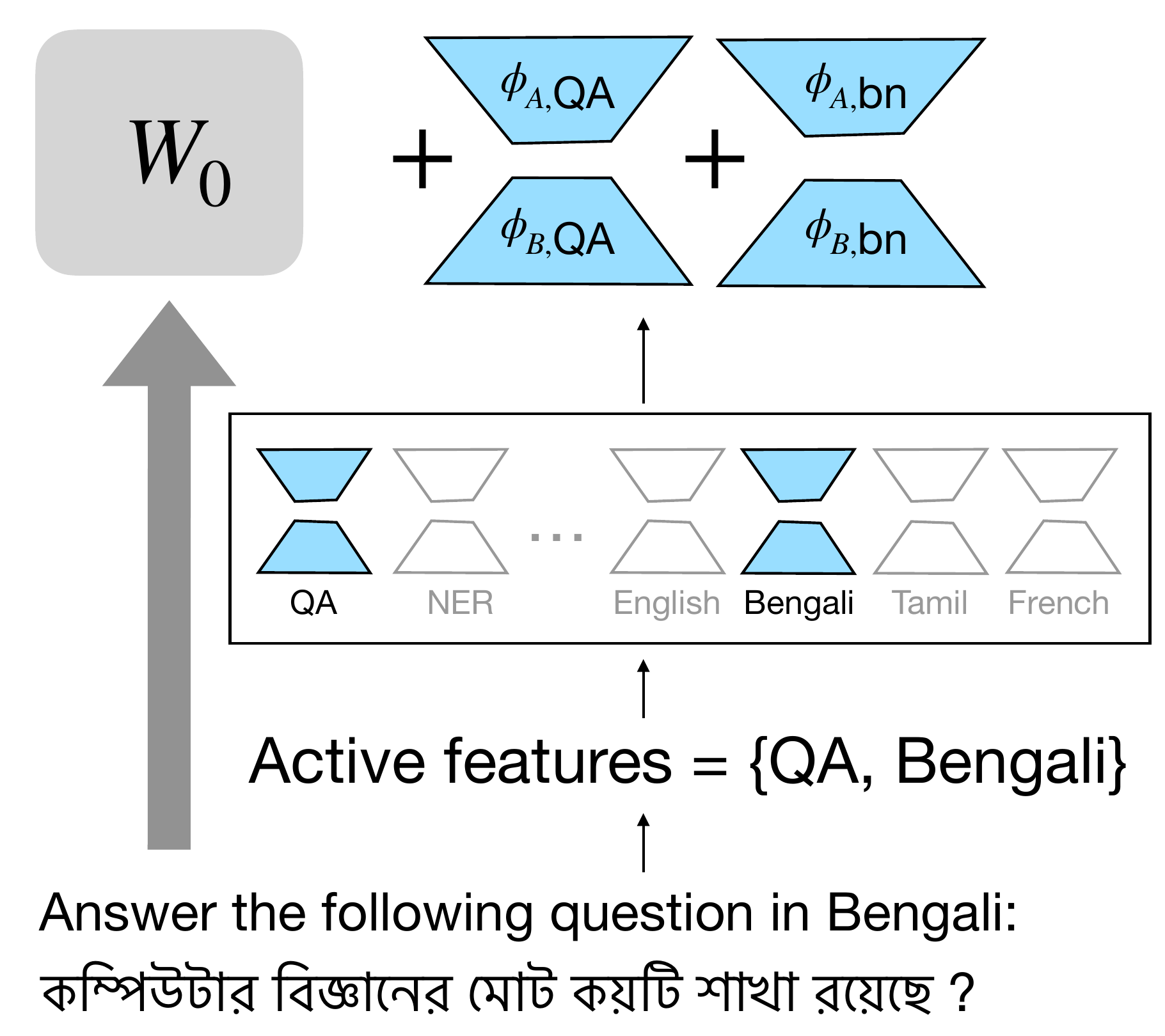}
    \caption{Training \model on diverse data mixtures.}
    \end{subfigure}
    \begin{subfigure}{0.5\textwidth}
    \centering
    \includegraphics[width=\textwidth]{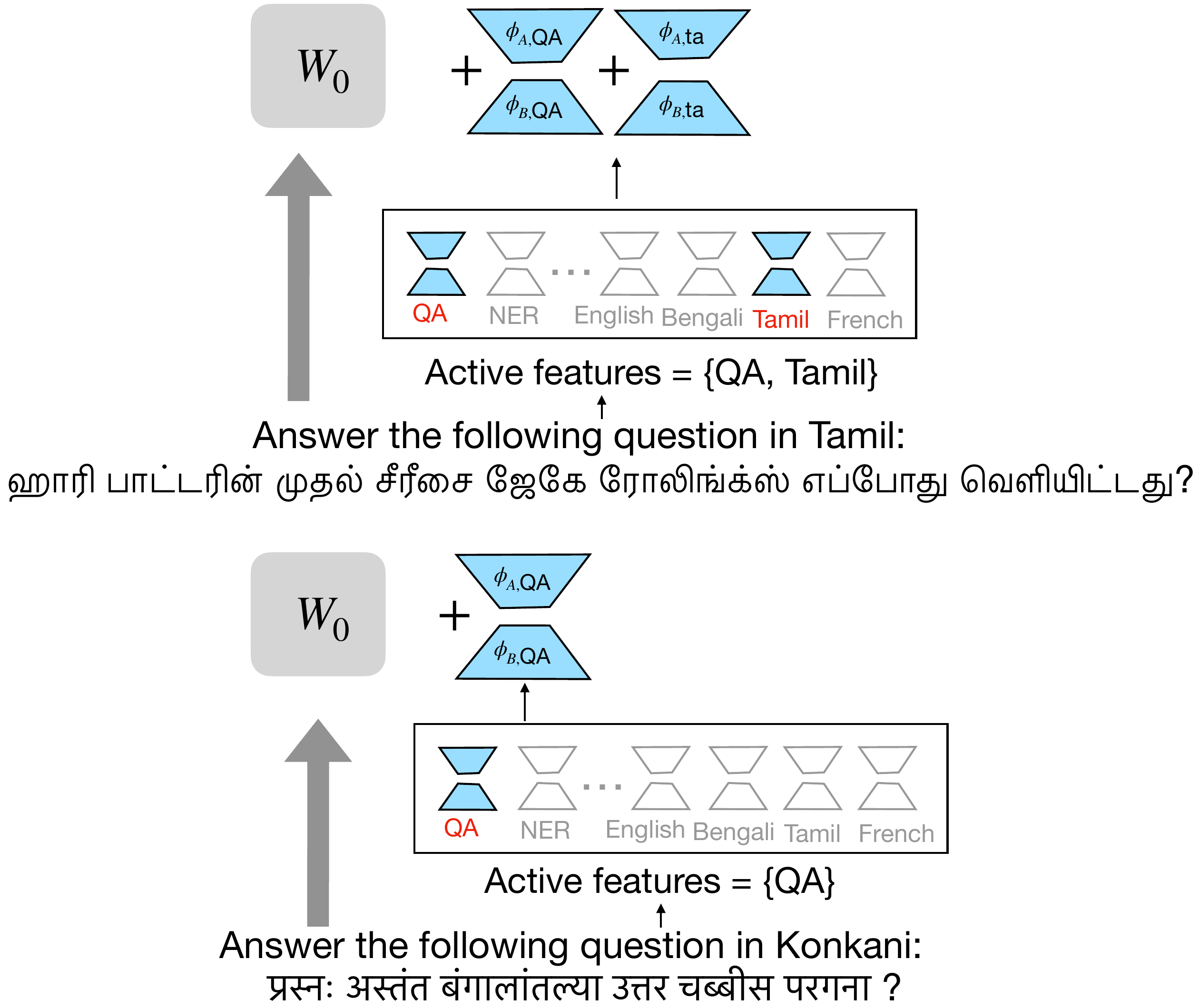}
    \caption{Inference with trained \model. \textit{Top}: \model naturally supports new test data with feature combinations unseen at training time; \textit{Bottom}: \model also supports new test data with partially unseen features.
    Since the Konkani language was not in the training mixture, \model only uses the available parameters corresponding to the QA feature.}
    \end{subfigure}
    \caption{Training and inference with \model.}
    \label{fig:flora_train_infer}
\end{figure*}

\section{Multitask multilingual learning}
\label{sec:background}

We consider a multitask learning \citep{Caruana1997MultitaskL} setting where datasets are indexed using task-language tuples.
We assume that there are $N$ tasks $\{w_1 \ldots w_N \}$ and $M$ languages $\{\ell_1 \ldots \ell_M \}$. And our goal is to serve fine-tuned LLMs for each of these $N \cdot M$ task-language combinations. It is worth noting that two special cases of this setup are multilingual modeling ($N=1$) and multitask modeling ($M=1$).

Parameter-efficient tuning~(PEFT) is a popular method to adapt a pretrained LLM to downstream tasks without incurring large computational cost. In this work, we want to use PEFT to support $O(M \cdot N)$ languages and tasks.

\subsection{Low-rank Adaptation}

\label{sec:background-lora}
We focus on Low-rank Adaptation~(LoRA)~\citep{hu2022lora}, an effective PEFT method that incurs minimal additional inference cost.  
Let $W_0 \in \Real^{d \times k}$ be the pretrained weight matrix, LoRA keeps the much larger weight matrix $W_0$ unchanged, and instead only optimizes low-rank factorized weight adaptation matrix $\Delta W = {\phi}^A {\phi}^B$. The final weight matrix is
\begin{align}
\label{eqn:lora}
    W = W_0 + \Delta W = W_0 + {\phi}^A {\phi}^B,
\end{align}
where ${\phi}^A \in \Real^{d \times r}$, ${\phi}^B \in \Real^{r \times k}$. Empirically LoRA often compares competitively with full fine-tuning, even when $r \ll \min(d, k)$. LoRA can thus significantly reduce the size of trainable parameters.

\subsection{Challenges}
\label{sec:peft-challenges}
PEFT methods such as LoRA have been shown to be very effective for tuning LLMs, achieving comparable results to full fine-tuning while incurring only a fraction of the computational cost.
However, there are several challenges with the multitask multilingual learning problem that the current PEFT methods might not be able to address.

\paragraph{Interference among different datasets}
Multitask multilingual training with PEFT requires one to fine-tune an LLM on a diverse data mixture, from up to $N \cdot M$ different datasets spanning $N$ tasks and $M$ languages. This approach has significantly lower overhead than modeling each dataset individually, and allows positive transfer. However, training a single set of PEFT parameters over all datasets could also lead to negative interference among tasks and languages that are dissimilar.

\paragraph{Generalizing to unseen task and language}
Publicly available wide-coverage multitask and multilingual datasets are often incomplete. For example, many task-language combinations are missing in XTREME-UP. Moreover, some underrepresented languages may still be missing from such datasets.
Standard PEFT methods could have difficulty generalizing to unseen task-language combinations and unseen languages: they simply optimize a single set of parameters on all tasks without explicit modeling assumptions that capture the relationships and similarities between different datasets. And it is not clear how to transform such PEFT parameters to a new task or language at inference time.

%% file: method_v2.tex
\section{Featurized Low-Rank Mixtures}

\label{sec:modularized-multitask-multilingual-lora}

We propose Featurized Low-Rank Mixtures~(\model), an effective multitask PEFT method that supports diverse training data mixture, and excels at zero-shot generalization to new task-language combinations.
Under \model, NLP tasks and languages are featurized as discrete features. And each feature is associated with a low-rank weight update matrix.  \cref{fig:flora_train_infer} shows the training and inference processes of \model.

\subsection{Model Architecture}

Given a diverse data mixture from $N$ tasks where each tasks are in $M$ languages, first we define a set of $D = N+M$ features where each feature could represent either a task, a language, or any other data attribute. We assign a low-rank factorized weight matrix ${\phi}^A_{i} {\phi}^B_{i}$ for each feature $i \in [1, D]$. 

Let $\vecc{x}$ be an input to the model, let $f(\vecc{x})$ represent the features of $\vecc{x}$, where  $f_i(\vecc{x}) = 1$ indicates that $\vecc{x}$ has feature $i$, and  $f_i(\vecc{x}) = 0$ otherwise.
\begin{align}
    W(\vecc{x}) &= W_0 +  \sum_{ i = 1 }^{D}  {f_i(\vecc{x})} {\phi}^A_{i} {\phi}^B_{i},
    \label{eqn:modularized_lora}
\end{align}
where ${\phi}^A_{i} \in \Real^{d \times r_i}$, ${\phi}^B_{i} \in \Real^{r_i \times k}$, and $r_i$ is the maximum rank of the the $i$-th feature's adaptation matrix.
Note that compared to LoRA in \cref{eqn:lora} that applies the same $\Delta W$ (and therefore same $W$) for all inputs, \model uses different adaptation matrices based on the features of the input data $f(\vecc{x})$.

\paragraph{Feature dropout.}
One potential problem of training \model is that the model might become overly reliant on the feature annotation of the training dataset, limiting positive transfer and making the model brittle to inputs different from training distribution.
We randomly turn off a subset of active features at training time with a predetermined feature dropout probability.
Experiments in \cref{sec:ablations} show that feature dropout brings consistent gains to \model. 

\paragraph{Exploiting feature sparsity for low training and serving costs.} 
Note that \cref{eqn:modularized_lora} implies that on input $\vecc{x}$, $W(\vecc{x})$ is not a function of $\{ \phi^A_i, \phi^B_i \}$ if $f_i(\vecc{x}) = 0$. In other words, weights of unused features of input $\vecc{x}$ are not needed in either training or serving time.   While the number of trainable parameters under \model grows linearly with the number of features $D$, the feature count of each input $\vecc{x}$ remains a constant; and this is a relatively small value in our multitask multilingual learning settings. Therefore, the compute costs of \model could still remain constant when scaling to increasingly more tasks and languages.

\subsection{Zero-shot Composition}
\label{sec:method_zeroshot}
We find \model to be particularly effective at zero-shot generalization, likely because of the explicit feature composition assumption. While previous work proposed using language-specific modules at \textit{pretraining} time to enhance crosslingual transfer \citep{vu2022overcoming,pfeiffer2023mmt5}, our work shows that sparse modularized PEFT architecture is also effective for directly adapting \textit{dense} pretrained models to unseen datasets.  

In this paper, we consider how \model adapts to unseen datasets in two different zero-shot scenarios:
\begin{description}
\item[Unseen combinations.] We want to do inference for a dataset that has a combination of active features that did not appear in the training mixture. For example, say the training data mixture contains QA data in French and semantic parsing data in Bengali; and we want to test the model on QA in Bengali. \model naturally supports such features; and no change is required while applying \cref{eqn:modularized_lora}.
\item[Unseen languages.] The test data could have a subset of features that are not covered by any of the dataset in the training mixture. Specifically, we focus on the setting where the test data is from a new language unseen during training. In this case, we only use the task feature of the data to calculate the model weights in \cref{eqn:modularized_lora}.
\end{description}
In \cref{sec:experiment-results}, we show that \model significantly outperforms the baselines for both types of zero-shot settings.

%% file: experiment.tex
\section{Experiments}
\label{sec:experiment-setup}

For all experiments, we use instruction fine-tuned PaLM-2 small \cite{anil2023palm} as the base model. We evaluate our method on a variety of tasks, languages, and data mixtures to verify that it generalizes to different use cases.

\paragraph{Datasets.}
We use the XTREME-UP dataset \citep{ruder2023xtremeup} and format the input data with the language and task information~\citep{wang2023fiat}.
We use different language and task subsets from XTREME-UP. Dataset details are described in \cref{sec:experiment-results}.\footnote{A list of all experiments' training and evaluation tasks can also be found in \cref{sec:train-eval-splits}.}

\paragraph{Metrics.} We report F1 scores for both in- and cross-lingual QA tasks, and NER tasks. For semantic parsing, we report exact-match accuracy. All numbers are normalized between $[0, 100]$. We report average normalized metric scores for (monolingual) multitask experiments.

\paragraph{Hyperparameters.} We use a batch size of $512$ during training for all experiments. For \model, we also set the feature dropout probability to be $0.7$.

\paragraph{Ranks.} Compute-matched LoRA baselines have rank-$6$ adaptation matrices\footnote{Strictly speaking their ranks $\leq 6$; we use this terminology loosely to reduce clutter.} in all experiments. In this paper, every dataset has both task and language features; and we allocate feature-specific parameter counts such that the trainable parameters of active features under \model always match compute-matched baselines. More specifically, we let task and language features have either rank-$2$ or rank-$4$ adaptation matrices. We allocate smaller matrices (rank-$2$) for task features in multitask experiments (\cref{sec:mt-or-ml}), and larger matrices in both multilingual (\cref{sec:mt-or-ml}) and joint multitask-multilingual (\cref{sec:joint-mtml}) experiments. We adjust the ranks of language features' adaptation matrices accordingly, to ensure that every dataset's adaptation matrices have a total rank $=6$.

\paragraph{Model selection.} We evaluate on validation splits every 200 iterations, and train for a maximum of 2000 steps. For every multilingual and multitask task, we choose the checkpoint that has the best averaged metric numbers across languages or tasks, and subsequently evaluate on the test splits.

\paragraph{Baselines.}
We compare to the vanilla LoRA method under several different settings to ensure the fairness of comparison:
\begin{itemize}
    \item \textbf{Compute-matched} sets the rank $r$ of the vanilla LoRA model to be equivalent to the maximum sum of the rank of feature-specific adaptations under \model. This ensures LoRA and \model uses comparable computation during training and inference.
    \item \textbf{Param-matched} sets the rank $r$ of the vanilla LoRA model such that the total number of trainable parameters is the same as its \model counterpart.\footnote{The number of trainable parameters in \model grows linearly with the number of features in a dataset. Therefore the param-matched counterpart's rank can be a relatively large number from around 20 to 90, depending on the task.} %
\end{itemize}

\section{Results}
\label{sec:experiment-results}

\subsection{Study A: Joint Multitask Multilingual Learning}
\label{sec:joint-mtml}

\begin{table*}
    \centering
    \resizebox{\textwidth}{!}{
    \begin{tabular}{c|r|r|r|r|r|r|rr}
    \toprule
     Method  & NER & Semantic Parsing & QA InLang  & QA CrossLang & \multicolumn{1}{|p{.18\linewidth}|}{Zero-shot Unseen Comb: QA CrossLang} & \multicolumn{1}{|p{.18\linewidth}|}{Zero-shot Unseen Comb: Semantic parsing}  & \multicolumn{1}{|p{.18\linewidth}|}{Zero-shot Unseen Lang: QA CrossLang} & Avg. \\
     \cmidrule(lr){1-1}  \cmidrule(lr){2-5} \cmidrule(lr){6-8} \cmidrule(lr){9-9}
      Compute-matched      & 76.3 & 35.4 & 86.6 & 83.1   & \num{82.11495209} & \num{17.7079703} & \num{74.56821738} & 65.1 \\ 
     Param-matched        & \num{81.64579773} & \textbf{47.1} & 87.3 & \num{83.18923696} & \num{80.02549057} & \num{28.43593619} & \num{74.71656842} & 68.9 \\ 
          \cmidrule(lr){1-1}  \cmidrule(lr){2-5} \cmidrule(lr){6-8} \cmidrule(lr){9-9}
     \model     & \textbf{84.0} & \num{45.59064865} & \textbf{89.4} & \textbf{85.2}  & \textbf{82.8} & \textbf{42.6} & \textbf{77.2} & \textbf{72.4} \\
    \bottomrule
    \caption{Multilingual multitask learning with \model (complete tables are in \cref{sec:joint-mtml-full}). Note that {\bf QA CrossLang} results in this table are evaluated only on XOR-TyDiQA languages, and are \emph{not} directly comparable against those from \cref{tab:single-feature}. \model achieves significant improvements over the compute-matched LoRA and it out-performs param-matched LoRA on three out of the four tasks. \model also achieves much better performance on zero-shot generalization in cross-lingual QA. The details of these experiments are discussed in \S\ref{sec:joint-mtml}.}
    \label{tab:multilingual-multitask-results}
    \end{tabular}
    }
\end{table*}

In this section, we evaluate the performance of \model and baselines on a diverse data mixture, where the tuning data contain different languages and tasks.

\subsubsection{Multitask multilingual tuning}
\label{sec:joint-mtml-supervised}
We conduct multitask multilingual tuning on both vanilla LoRA and our proposed \model model. This setting shows one of the primary strengths of \model: The ability to simultaneously generalize along multiple axes (tasks and languages) without suffering from negative task interference. We evaluate on four tasks from XTREME-UP covering a wide variety of use cases and languages: crosslingual QA, in-language QA, semantic parsing, NER. In addition, we also add training data from machine translation to the data mixture since it has the best language coverage which allows cross-lingual transfer.
We use all languages in in-language QA, semantic parsing, NER, and a subset of languages in cross-lingual QA as training data.\footnote{\label{ft:xor-tydiqa}We only use the languages included in the original XOR-TyDi QA~\citep{asai-etal-2021-xor}: Arabic, Bengali, Japanese, Finish, Korean, Russian, Telugu. The rest of the cross-lingual QA languages \Dash all Indic languages \Dash are evaluated for zero-shot generalization in \cref{sec:zero-shot-generalization}.} In addition, we also include the corresponding machine translation datasets of languages from these 4 datasets.

\paragraph{\model significantly outperforms baselines under multitask multilingual tuning.}
The overall results are listed in \cref{tab:multilingual-multitask-results}. We can see that \model out-performs the the best LoRA baseline for all tasks other than semantic parsing. While it loses to param-matched LoRA on semantic parsing, \model~has significantly less computational cost in comparison and it significantly out-performs the vanilla LoRA with the same computational cost.   

\subsubsection{Zero-shot Generalization}
\label{sec:zero-shot-generalization}
To evaluate both zero-shot generalizations (\cref{sec:method_zeroshot}) we prepare two different training datasets:
\begin{itemize}
\item \textbf{Holding out languages in cross-lingual QA.} We reuse the training dataset from the joint multitask multilingual learning setup (\cref{sec:joint-mtml}) to evaluate both unseen feature combinations and unseen languages in the cross-lingual QA task. For {\bf unseen combinations}, we evaluate the performance of the set of languages where the languages were present in other multilingual tasks in the training dataset; and for {\bf unseen languages}, we evaluate on the set of languages that do not appear in other multilingual tasks in the training dataset. 
\item \textbf{Holding out languages in semantic parsing.} In this scenario, we use portions of crosslingual QA, in-language QA, semantic parsing, NER, and machine translation datasets from XTREME-UP as our training dataset. We include full crosslingual QA, in-language QA, NER datasets; but we do not include underrepresented languages in the semantic parsing portion of the training data. We also only include machine translation portions of languages that are already available in the other $4$ subsets, as in \cref{sec:joint-mtml}.  And we evaluate the {\bf unseen combination} performance of all held-out languages under semantic parsing.\footnote{These held-out languages are unseen combinations, rather than unseen languages, since they are available in the machine translation datasets of XTREME-UP.}
\end{itemize}

\paragraph{Our method is very effective at both types of zero-shot generalization.} The comparison of \model and baseline methods can be found in the rightmost 3 columns in \cref{tab:multilingual-multitask-results}. We can see that \model brings significant improvements to both unseen combinations and unseen languages on both cross-lingual QA and semantic parsing. This is likely because \model allows one to select the subset of parameters most relevant to the features of the new test data, allowing more effective zero-shot generalization.  

\subsection{Study B: Multitask or multilingual learning}

\label{sec:mt-or-ml}
In this section, we examine the performance of our method and the baselines on data mixtures with a single type of feature. That is, we train and evaluate on datasets of a single task from several different languages~(multilingual learning), or datasets of a single language with different tasks~(multitask learning).  We also add additional baselines for this setting where we train a separate LoRA model for each individual dataset (denoted as Single-Lang and Single-Task in \cref{tab:single-feature}). This approach alleviates the capacity constraint of vanilla LoRA on diverse datasets, but adds more engineering overhead (as we briefly noted in \cref{sec:intro}) and cannot leverage positive transfer between datasets.

Specifically, we use subsets of the XTREME-UP dataset to construct these datasets:
\begin{itemize}
    \item \textbf{Multilingual learning.} We experiment on four tasks: NER, semantic parsing, in-language QA, and cross-lingual QA. For each task, we train and evaluate on all languages available in the XTREME-UP dataset. Unlike \S\ref{sec:joint-mtml}, we train a separate multilingual model for each task in this setting.
    \item \textbf{Multitask learning.} We evaluate on two under-represented languages: Swahili and Bengali. We use the subset of tasks available for each language in XTREME-UP. For Swahili, we use semantic parsing, QA (in-language), and NER. For Bengali we use semantic parsing and QA (both in-language and cross-lingual).\footnote{The mismatch between task choices between these two languages is due to the sparsity of available datasets in XTREME-UP.} Unlike \S\ref{sec:joint-mtml}, we train a separate multitask model for each language in this setting.
\end{itemize}

\begin{table*}
    \centering
    \resizebox{\textwidth}{!}{
    \begin{tabular}{crrrrr|rrr}
    \toprule
     \multirow{2}{*}{Method} & \multicolumn{4}{||c||}{Multilingual Learning} & \multicolumn{2}{|c|}{Multitask Learning} & \multirow{2}{*}{Avg.} \\
      \cmidrule(lr){2-5} \cmidrule(lr){6-7} 
       & NER & Semantic Parsing & QA InLang &  QA CrossLang  & Swahili & Bengali & \\
     \cmidrule(lr){1-1}  \cmidrule(lr){2-5} \cmidrule(lr){6-7} \cmidrule(lr){8-8}
    Compute-matched      & \num{81.29791379} & \num{40.49185181} & \num{88.46985287} &  \num{77.27444282}   & \num{66.04435356} & \num{63.83674049} & 69.6 \\
   Param-matched       & \num{83.22806269} & \textbf{47.2} & \num{89.0400806} &  \num{76.19151071}  & \num{66.77800091} & \num{65.78626569} & 71.3 \\
     \cmidrule(lr){1-1}  \cmidrule(lr){2-5} \cmidrule(lr){6-7} \cmidrule(lr){8-8}
  \model        & \textbf{84.3} & \num{45.03063347} & \textbf{89.4} &  \textbf{77.6}     & \textbf{70.2} & \textbf{69.0}  & \textbf{72.6} \\
    \bottomrule
    \end{tabular}
    }
    \caption{Results of either multitask learning or multilingual learning (complete tables are in \cref{sec:single-feature-full}). While there is not a single baseline that is consistently better than others, \model leads to consistent improvement over the LoRA baseline with similar computational cost. See \S\ref{sec:mt-or-ml} for a description of these settings.}
    \label{tab:single-feature}
\end{table*}

\paragraph{No baseline method consistently out-performs other baselines.}
We report the performance of our method and the baselines in \cref{tab:single-feature}. For each experiment, we list the average result over all languages or tasks in the datasets. First, we find that among the different baseline methods, there is no method that consistently out-performs others. Specifically, param-matched LoRA tends to have advantage for multitask learning and tasks that are very different from pretraining~(semantic parsing and NER), while compute-matched LoRA appears to be superior on multilingual QA tasks. We suspect that this is because param-matched LoRA benefits from its higher capacity for the model to learn to generate into targets that are very different from the pretraining data, and it is also helpful for supporting generation into diverse target tasks. However, param-matched LoRA has significantly higher numbers of trainable parameters. This could lead to much higher computational overhead compared to compute-matched LoRA. Moreover, we observe that compute-matched LoRA is actually more competitive on tasks such as crosslingual QA, likely because it reduces over-fitting by tuning a much smaller number of parameters.

\paragraph{\model achieves much better performance than vanilla LoRA with the same computation budgets.}
\model consistently outperforms compute-matched LoRA baseline on all settings, achieving significant gains without adding additional serving cost. Furthermore, our proposed method also outperforms param-matched LoRA on five out of the six data mixtures we evaluated. 

We hypothesize that param-matched LoRA is better than \model at semantic parsing but worse at other tasks because semantic parsing requires the LLM to generate into structured data that are very different from the pretraining data distribution, which might require particularly large model capacity to learn. In fact, \cref{tab:single-feature} shows that param-matched LoRA with a higher rank is much better than compute-matched LoRA for semantic parsing, while being worse or comparable on question answering tasks. While param-matched LoRA could be particularly helpful for the semantic parsing task, it requires more computational resources to scale to large number of datasets. These results indicate that our method is an effective and computationally-efficient strategy for tuning LLMs on diverse data mixtures.

%% file: ablations.tex
\section{Analysis and Ablations}
\label{sec:ablations}

\subsection{Effect of feature dropout}

\begin{table}[ht]
\begin{tabular}{lll}
\toprule
 Feature dropout strength     & Validation        & Test              \\ \cmidrule(lr){1-1} \cmidrule(lr){2-3}
$p=0.7$ & \num{87.57202827} & \num{89.29366896} \\
$p=0.5$ & \num{87.31078593} ($-0.3$) & \num{88.79437595} ($-0.5$) \\
$p=0.3$ & \num{87.10642158} ($-0.5$) & \num{88.66860453} ($-0.6$) \\
$p=0.0$ & \num{86.89309438} ($-0.7$) & \num{88.92305586} ($-0.4$) \\
\bottomrule
\end{tabular}
\caption{Effects of feature dropout on in-language QA. Removing feature dropout leads to consistent performance drop.}
\label{tab:reg-cqa}
\end{table}

In this section, we evaluate the effectiveness of feature dropout for \model. We compare the performance of \model with and without feature dropout for in-language QA task using multilingual training, and the results are in \cref{tab:reg-cqa}. We can see that removing feature dropout leads to significant performance drop for both validation and test set. We hypothesize that this is because feature dropout is an effective regularization that encourages \model to utilize the shared parameters for positive transfer between tasks and languages.

\begin{table*}[ht]
    \centering
    \begin{tabular}{lllll}
    \toprule
      Rank & QA CrossLang & QA InLang & Semantic Parsing & NER  \\
     \cmidrule(lr){1-1} \cmidrule(lr){2-5}
     $2$   & 77.6  & 89.4  & 45.0  & 84.3   \\
     $1$   & 76.5~($-1.1$)  & 89.3~($-0.1$)  & 45.2~($+0.2$)  & 83.1~($-1.2$)   \\
    \bottomrule
    \end{tabular}
    \caption{Effect of different ranks for language-specific adaptation matrices in \model with multitask multilingual tuning. Decreasing the rank of matrices leads to very small drop in performance on most tasks, but it has a large negative effect on semantic parsing.}
    \label{tab:rank_diff}
\end{table*}

\subsection{Effect of rank for \model}
\label{subsec:reduce_rank}
In multilingual experiments (\cref{tab:single-feature}) the language features only have rank-$2$ feature-specific weight update matrices.
Such low rank configurations help reduce the overall parameter count.
In this section, we examine how well \model performs under an even smaller budget. \cref{tab:rank_diff} shows the results of \model on multitask multilingual tuning with rank set to 4 and 2. We can see that reducing the capacity of the feature-specific weight update matrices actually only lead to a small drop in performance on most tasks, indicating that \model could be effective under even restrictive computational requirements.

\subsection{\model performs increasingly better on a diverse training dataset}
\begin{figure}
    \centering
    \includegraphics[width=0.6\textwidth]{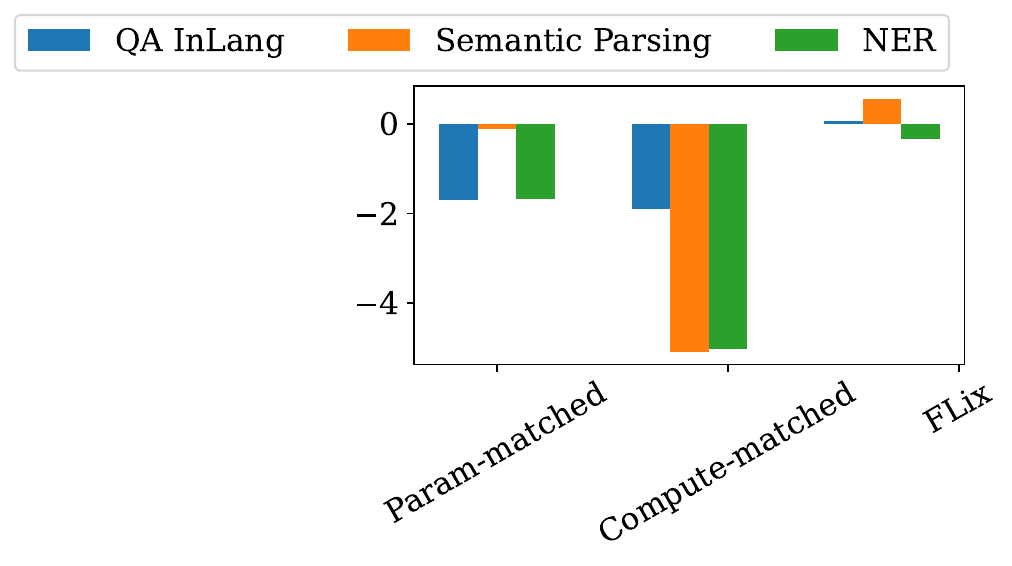}
    \caption{Difference in performance between multitask multilingual tuning and multilingual tuning only. While using the more diverse multilingual multitask data mixture leads to large performance drop for vanilla LoRA methods, \model generally maintains or slightly improves the task performance with more diverse data mixtures.}
    \label{fig:diverse_data_compare}
\end{figure}

In \cref{sec:experiment-results} we examined the performance of \model and baselines under each data mixtures. Here we want to compare how different methods perform when using increasingly more diverse data mixtures. Specifically, we compare the change in task performance when using the multitask multilingual mixture as opposed to using only the multilingual data for each task. \cref{fig:diverse_data_compare} shows the results of \model and the baselines. We can see that vanilla LoRA suffers from negative transfer \cite{Wang2018CharacterizingAA}. In particular, the compute-matched version with a smaller rank has significant decrease in performance on a diverse multitask multilingual training dataset.  On the other hand, \model is able to generally maintain similar or slightly higher performance when using the more diverse data mixture. This result indicates that \model is a superior PEFT method to scale to large number of tasks and languages.

%% file: related_work.tex
\section{Related Work}

\paragraph{Multilingual/multitask PEFT}
While most prior works on parameter-efficient tuning focus on tuning the model for a single task, some recent works propose post-hoc composition pretrained PEFT modules to generalize to new task or language~\citep{pfeiffer-etal-2020-mad,huang2023lorahub,wang-etal-2021-efficient-test,chronopoulou2023language}. These methods generally require separate training runs for each dataset, while our work propose a parameter-efficient tuning method that adapts the LLM using diverse data mixture in a single training run. \citet{vu2022overcoming} proposed to add a multilingual pretraining stage to prompt tuning, which shows some improvements for zero-shot generalization when adapting to cross-lingual summarization task. In comparison, our proposed \model focuses on tuning using multilingual data in many downstream tasks without additional training on unlabeled pretrainining data.  \citet{wang2023multitask} proposed a multitask prompt tuning method to learn a single prompt which could be used to adapt to many other target tasks. Similarly, this method requires multiple training stage while our method allows training end-to-end. Both of these methods are built upon prompt tuning using either multilingual training in a single task or multitask tuning in English, while our method supports diverse datasets in different tasks languages or any other arbitrary features.      

\paragraph{Mixture-of-experts Models}
Mixture-of-experts models (MoEs)~\citep{shazeer2017outrageously,lepikhin2020gshard,jawahar-etal-2023-automoe} are effective at increasing the model capacity by adding multiple expert parameters which could be activated differently to support different inputs. This architecture has been used to improve both pretrained models~\citep{lepikhin2020gshard,jawahar-etal-2023-automoe} and parameter-efficient tuning methods~\citep{zadouri2023loramoe, zhu2023sira}. Since MoEs often adds more computational cost, many works try to reduce the cost and improve the effectiveness of the model by either task-level routing (\emph{e.g.}, Task MoEs proposed by \citet{kudugunta-etal-2021-beyond-distillation}) or encouraging sparsity of the experts~\citep{shazeer2017outrageously,lepikhin2020gshard}. \model resembles Task MoEs in that \model leverages task information as well. But \model has additional composition capabilities thanks to its featurization. %

\paragraph{Modularized Pretrained Models}
Previous work proposed to add language-specific parameters to multilingual pretrained models~\citep{pfeiffer-etal-2022-lifting,pfeiffer2023mmt5} or machine translation models~\citep{zhang-2020-share-mnmt}. These works showed that language-specific modules often bring significant improvements to multilingual tasks. And they are especially helpful for zero-shot cross-lingual transfer. While prior works added language-specific modules during multilingual pretraining, 
in this paper we focus on the problem of adapting a pretrained model to a diverse mixture with many tasks and languages.

%% file: conclusion.tex
\section{Future Work}
There are several promising future directions for our work. While \model achieves good zero-shot performance, it could potentially benefit from methods that automatically learn to select and compose parameters pretrained on different features for unseen data. It is also interesting to examine other applications of \model: we encoded task and language informations as features in this work. But potentially other properties, such as modality, could also be featurized under \model. 

\section{Conclusion}
In this paper, we propose Featurized Low-Rank Mixtures~(\model), an effective parameter-efficient tuning method to fine-tune pretrained LLMs to data mixtures containing datasets in diverse tasks and languages. Our experiments show that \model leads to significantly better performance on multitask multilingual fine-tuning compared to standard LoRA with little computational overhead. We also find that \model achieves much better zero-shot generalization to new languages and task language combination unseen at training time.

%% file: appendix.tex
\section{Constant parameter sharing hurts \model}

\label{sec:effects-task-specific-features}

\begin{table*}
\begin{tabular}{lllll}
\toprule
Configuration & QA InLang & QA CrossLang & Semantic parsing & NER \\ 
\cmidrule(lr){1-1} \cmidrule(lr){2-5}
 \model & \num{89.44300842} & \num{85.2} & \num{45.59064865} & \num{84.0}  \\
 + shared parameters & \num{88.23865085} & \num{83.97255834} & \num{41.89471944} & \num{80.67657799} \\ 
\cmidrule(lr){1-1} \cmidrule(lr){2-5}
 Compute-matched LoRA & $86.6$ & $83.1$ & $35.4$ & $76.3$ \\
\bottomrule
\end{tabular}
\caption{Effects of adding shared parameters across all datasets for \model on multitask multilingual learning. Adding shared parameters leads to decreased performance for \model. Yet the presence of task and language features still provides significant improvement over compute-matched LoRA. 
}
\label{tab:globally-shared-features}
\end{table*}

Our proposed \model routes each input to their feature-specific adaptations, based on the dataset features.
While constant parameter sharing happens when some features are always active for all datasets in a mixture (\emph{e.g.}, in experiments in \cref{sec:mt-or-ml}), this is not generally true.
In this section, we look into the effects of enforcing constant parameter-sharing in  \model. Specifically, we design a dummy feature that is always active to all datasets in the joint multitask multilingual training mixture used in \cref{sec:joint-mtml-supervised}, which implies constant parameter sharing.
We then compare \model models both without and with this dummy feature, and also with a vanilla LoRA baseline, all under comparable compute budgets.\footnote{Specifically, we rearrange the rank sizes of all features such that the sum is $6$. The dummy feature has rank$=4$, and the language and task features have rank$=1$. The compute-matched LoRA baseline does not make use of dataset features, and has rank$=6$.}

In \cref{tab:globally-shared-features}, we can see that adding a shared parameter actually leads to worse performance for \model with similar computational cost. However, \model with constantly shared features still outperforms computed-matched LoRA, which does not leverage task or language features at all.

\section{Training and evaluation datasets}
\label{sec:train-eval-splits}

\subsection{Multilingual learning}
\label{sec:multilingual-splits}
The (mono-task) multilingual learning experiments described in \cref{sec:mt-or-ml} train and evaluate on the same languages. Their language and locale codes are:
\begin{description}
\item[Semantic parsing] am, be, bn, fi, ha, hu, ja, pt\_br,	ru,	sw,	ta,	tr,	yo,	zu,	de\_localized,	en,	de,	es,	fr,	hi,	th
\item[In-language QA] ar,	bn,	fi,	id,	ko,	ru,	sw,	te,	en
\item[Cross-lingual QA] ar,	bn,	fi,	ko,	ru,	te,	as,	bho,	brx,	gbm,	gom,	gu,	hi,	hne,	kn,	mai,	ml,	mni,	mr,	mwr,	or,	pa,	ps,	sa,	ta,	ur
\item[NER] am,	bm,	bbj,	ee,	ha,	ig,	rw,	lg,	luo,	mos,	ny,	pcm,	sn,	sw,	tn,	tw,	wo,	xh,	yo,	zu
\end{description}

\subsection{Multitask learning}
The (mono-lingual) multitask learning experiments described in \cref{sec:mt-or-ml} train and evaluate on $3$ different tasks, for the $2$ languages we evaluate on respectively. They are:
\begin{description}
\item[Swahili (sw)] Semantic parsing, NER, in-language QA
\item[Bengali (bn)] Semantic parsing, in-language QA, cross-lingual QA
\end{description}

\subsection{Joint multitask multilingual learning}
\subsubsection{Training}
\label{sec:joint-mt-ml-training}
The joint multitask multilingual learning experiments in \cref{sec:joint-mtml} use the union of the following datasets:
\begin{description}
\item[Semantic parsing] am, be, bn, fi, ha, hu, ja, pt\_br,	ru,	sw,	ta,	tr,	yo,	zu,	de\_localized,	en,	de,	es,	fr,	hi,	th
\item[In-language QA] ar,	bn,	fi,	id,	ko,	ru,	sw,	te,	en
\item[Cross-lingual QA] ar, bn, fi, ko, ru, te
\item[NER] am,	bm,	bbj,	ee,	ha,	ig,	rw,	lg,	luo,	mos,	ny,	pcm,	sn,	sw,	tn,	tw,	wo,	xh,	yo,	zu
\item[Machine translation] id, es, hi, yo, ja, lg, ny, ru, be, ar, de, bn, fr, tr, ig, th, fi, zu, te, ko, sw, xh, hu, ha, sn, ta, am
\end{description}
\subsubsection{Multilingual evaluation}
We evaluate models trained on the dataset described in \cref{sec:joint-mt-ml-training} on the $4$ multilingual subsets of the training dataset. The results are reported in the first $4$ columns of \cref{tab:multilingual-multitask-results}.
\subsection{Zero-shot generalization}
\subsubsection{Training}
For zero-shot evaluations on cross-lingual QA datasets in \cref{sec:zero-shot-generalization}, we reuse the joint multitask multilingual training dataset described in \cref{sec:joint-mt-ml-training}.
For evaluation of zero-shot unseen combinations on semantic parsing datasets in the same section, we use the union of the following datasets:
\begin{description}
\item[Semantic parsing] fi, hu, ja, pt\_br,	ru,	tr,	de\_localized,	en,	de,	es,	fr,	hi
\item[In-language QA] ar,	bn,	fi,	id,	ko,	ru,	sw,	te,	en
\item[Cross-lingual QA] ar,	bn,	fi,	ko,	ru,	te,	as,	bho,	brx,	gbm,	gom,	gu,	hi,	hne,	kn,	mai,	ml,	mni,	mr,	mwr,	or,	pa,	ps,	sa,	ta,	ur
\item[NER] am,	bm,	bbj,	ee,	ha,	ig,	rw,	lg,	luo,	mos,	ny,	pcm,	sn,	sw,	tn,	tw,	wo,	xh,	yo,	zu
\item[Machine translation] de, mni, mr, ig, id, mwr, ko, fi, ta, bn, ar, es, ja, zu, be, gbm, tr, as, bho, yo, ml, sw, hi, am, bbj, lg, pcm, ny, tw, hu, luo, rw, brx, pt\_br, gu, or, gom, te, sn, wo, fr, ps, ha, hne, xh, en, sa, tn, de\_localized, ur, bm, ee, th, ru, pa, kn, mai, mos
\end{description}

\subsubsection{Zero-shot unseen combinations in cross-lingual QA}
\label{sec:zero-shot-unseen-comb-cqa}
In the `Zero-shot Unseen Comb: QA CrossLang' column of \cref{tab:multilingual-multitask-results} we report the average cross-lingual QA F1 scores of languages that are missing from the cross-lingual QA portion of the joint multitask multilingual training set  (\cref{sec:joint-mt-ml-training}), but present in other datasets of the same training set. The language and locale codes of these languages are: hi, ta.

\subsubsection{Zero-shot unseen languages in cross-lingual QA}
\label{sec:zero-shot-unseen-lang-cqa}
In the `Zero-shot Unseen Lang' column of \cref{tab:multilingual-multitask-results} we report the average cross-lingual QA F1 scores of languages that are missing from the joint multitask multilingual training set. The language and locale codes of these languages are: bho,	brx,	gbm,	gom,	hne,	mai,	mni,	mr,	mwr,	sa,	as,	gu,	kn,	ml,	or,	pa,	ps,	ur.

\subsubsection{Zero-shot unseen combinations in semantic parsing}
\label{sec:zero-shot-unseen-comb-semparse}
In the `Zero-shot Unseen Comb: Semantic parsing' column of \cref{tab:multilingual-multitask-results} we report the average exact match accuracy scores of languages that are designated as underrepresented in the XTREME-UP dataset, and held out from the training set in this experiment. These languages however appear in the machine translation portion of the training dataset. The language and locale codes of these languages are: am,	be,	bn,	ha,	sw,	ta,	yo,	zu,	th.

\clearpage

\section{Full results of \cref{tab:single-feature}}

\label{sec:single-feature-full}

In addition to the test results reported in \cref{tab:single-feature}, we include validation results as well.

\subsection{Multilingual experiments}
\subsubsection{Cross-lingual QA}
\begin{table}
\pgfplotstableread{
Language ar	bn	fi	ko	ru	te	as	bho	brx	gbm	gom	gu	hi	hne	kn	mai	ml	mni	mr	mwr	or	pa	ps	sa	ta	ur	Average
Validation 81.42288971	77.85762787	81.23019409	84.9673996	77.98108673	79.82839966	78.49246216	74.64076996	54.44968414	74.1272049	78.33480835	77.88486481	82.16366577	74.13703156	79.18672943	77.60512543	77.01988983	63.35907364	76.12967682	75.95301819	77.38455963	77.96183014	76.49067688	74.8692627	78.42307281	76.84483337	76.49022454
Test 80.79248047	83.67495728	79.75996399	83.93672943	80.4642334	81.52301788	78.72525787	75.76106262	47.38778687	73.37224579	75.77420807	80.23872375	83.54368591	77.05497742	80.05040741	76.95799255	79.27305603	63.10293579	78.0782547	75.92770386	77.6730957	79.82538605	77.44564056	78.27762604	78.50572968	76.66133881	77.06878838
}\singletaskcrossqatable
\pgfplotstabletranspose[string type, colnames from=Language, input colnames to=Language]{\transposedsingletaskcrossqatable}{\singletaskcrossqatable}
\pgfplotstabletypeset[columns/Language/.style={string type, column name={Language / Locale ID}},columns/Validation/.style={numeric type,precision=1,zerofill},columns/Test/.style={numeric type,precision=1,zerofill},
  every head row/.style={
    before row={\toprule},
    after row={\midrule}
  },
  every last row/.style={before row={\midrule}, after row=\bottomrule}
]{\transposedsingletaskcrossqatable}
\caption{Single-language compute-matched cross-lingual QA results.}
\label{tab:single-lang-compute-matched-cqa}
\end{table}

\begin{table}
\pgfplotstableread{
Language ar	bn	fi	ko	ru	te	as	bho	brx	gbm	gom	gu	hi	hne	kn	mai	ml	mni	mr	mwr	or	pa	ps	sa	ta	ur	Average
Validation 82.43021393	78.91338348	82.7832489	85.2975235	79.56187439	81.76415253	80.73427582	79.07290649	54.80800629	78.39836884	78.59101105	79.47558594	82.9749527	78.82534027	80.03065491	78.97795105	79.6931076	60.3968277	77.97235107	77.34370422	79.1818924	79.0575943	78.04851532	76.95947266	79.58598328	78.52624512	78.05404399
Test 83.43112183	83.30486298	82.64983368	88.31969452	83.15860748	81.49655151	78.9444046	75.49677277	47.51170731	73.33868408	76.80464935	78.77664948	83.68528748	77.8529129	79.86512756	75.60307312	78.95450592	60.47898483	77.35700989	75.60457611	78.06960297	78.56925201	77.93649292	77.16480255	78.68141174	76.07893372	77.27444282
}\singletaskcrossqatable
\pgfplotstabletranspose[string type, colnames from=Language, input colnames to=Language]{\transposedsingletaskcrossqatable}{\singletaskcrossqatable}
\pgfplotstabletypeset[columns/Language/.style={string type, column name={Language / Locale ID}},columns/Validation/.style={numeric type,precision=1,zerofill},columns/Test/.style={numeric type,precision=1,zerofill},
  every head row/.style={
    before row={\toprule},
    after row={\midrule}
  },
  every last row/.style={before row={\midrule}, after row=\bottomrule}
]{\transposedsingletaskcrossqatable}
\caption{Multiple-language compute-matched cross-lingual QA results.}
\label{tab:multiple-lang-compute-matched-cqa}
\end{table}

\begin{table}
\pgfplotstableread{
Language ar	bn	fi	ko	ru	te	as	bho	brx	gbm	gom	gu	hi	hne	kn	mai	ml	mni	mr	mwr	or	pa	ps	sa	ta	ur	Average
Validation 81.31397247	79.74903107	80.73872375	84.77822876	78.59945679	79.82579041	79.09648132	77.69944	58.14001465	76.88051605	76.92139435	78.24692535	84.1889801	78.07463074	79.71773529	80.30090332	79.72872925	63.35530853	78.54998016	77.25248718	76.27497101	77.62625122	74.75119019	78.77318573	80.34363556	78.90473175	77.68587288
Test 82.84125519	83.52205658	81.79576111	86.99436188	81.97655487	82.18575287	78.3418808	75.21755981	51.85730743	73.21691132	71.5278244	76.62319183	83.2144928	75.46587372	77.71244049	75.7896347	77.73670959	59.4626236	76.9765625	75.86991119	73.93021393	76.36155701	71.74909973	76.61592865	78.46063232	75.53318024	76.19151071
}\singletaskcrossqatable
\pgfplotstabletranspose[string type, colnames from=Language, input colnames to=Language]{\transposedsingletaskcrossqatable}{\singletaskcrossqatable}
\pgfplotstabletypeset[columns/Language/.style={string type, column name={Language / Locale ID}},columns/Validation/.style={numeric type,precision=1,zerofill},columns/Test/.style={numeric type,precision=1,zerofill},
  every head row/.style={
    before row={\toprule},
    after row={\midrule}
  },
  every last row/.style={before row={\midrule}, after row=\bottomrule}
]{\transposedsingletaskcrossqatable}
\caption{Multiple-language parameter-matched cross-lingual QA results.}
\label{tab:multiple-lang-param-matched-cqa}
\end{table}

\begin{table}
\pgfplotstableread{
Language ar	bn	fi	ko	ru	te	as	bho	brx	gbm	gom	gu	hi	hne	kn	mai	ml	mni	mr	mwr	or	pa	ps	sa	ta	ur	Average
Validation 83.0259552	79.15621185	82.71433258	85.87915039	79.55477142	79.9183197	81.71894836	78.84030151	56.86683273	78.14399719	78.37062836	79.7123642	82.23831177	78.1284256	81.22496033	78.70188141	80.433815	62.54589081	78.26036072	76.30103302	79.49866486	80.23826599	78.59462738	77.8562851	80.81924438	79.53885651	78.39547832
Test 83.55679321	84.56516266	83.39576721	87.44548035	84.34780884	80.28464508	77.75946808	76.34103394	50.64892197	73.59212494	76.33291626	77.09381866	83.60315704	76.25943756	81.01939392	77.60903931	78.73603821	63.43293381	77.73236847	74.84450531	78.11309052	78.15808868	77.01473236	78.02135468	78.79354858	77.61238861	77.55053916
}\singletaskcrossqatable
\pgfplotstabletranspose[string type, colnames from=Language, input colnames to=Language]{\transposedsingletaskcrossqatable}{\singletaskcrossqatable}
\pgfplotstabletypeset[columns/Language/.style={string type, column name={Language / Locale ID}},columns/Validation/.style={numeric type,precision=1,zerofill},columns/Test/.style={numeric type,precision=1,zerofill},
  every head row/.style={
    before row={\toprule},
    after row={\midrule}
  },
  every last row/.style={before row={\midrule}, after row=\bottomrule}
]{\transposedsingletaskcrossqatable}
\caption{\model cross-lingual QA results.}
\label{tab:multiple-lang-flora-cqa}
\end{table}

\clearpage

\subsubsection{In-language QA}

\begin{table}
\pgfplotstableread{
Language ar	bn	fi	id	ko	ru	sw	te	en	Average
Validation 87.27390289	85.78749084	89.45069885	85.32976532	81.65519714	87.01	84.32521057	90.365448	85.3263855	86.28045546
Test 88.45323181	86.35943604	90.17812347	86.03755951	84.63900757	85.08	87.54366302	92.3807373	87.14355469	87.53503482
}\singletaskcrossqatable
\pgfplotstabletranspose[string type, colnames from=Language, input colnames to=Language]{\transposedsingletaskcrossqatable}{\singletaskcrossqatable}
\pgfplotstabletypeset[columns/Language/.style={string type, column name={Language / Locale ID}},columns/Validation/.style={numeric type,precision=1,zerofill},columns/Test/.style={numeric type,precision=1,zerofill},
  every head row/.style={
    before row={\toprule},
    after row={\midrule}
  },
  every last row/.style={before row={\midrule}, after row=\bottomrule}
]{\transposedsingletaskcrossqatable}
\caption{Single-language compute-matched in-language QA results.}
\label{tab:single-lang-compute-matched-iqa}
\end{table}

\begin{table}
\pgfplotstableread{
Language ar	bn	fi	id	ko	ru	sw	te	en	Average
Validation 86.36549377	89.02135468	88.93639374	87.78442383	83.41203308	86.63763428	86.56314087	91.11143494	85.26934814	87.23347304
Test 87.60466003	85.90687561	89.08202362	88.73737335	86.22928619	87.77175903	89.5610199	93.08478546	88.25089264	88.46985287
}\singletaskcrossqatable
\pgfplotstabletranspose[string type, colnames from=Language, input colnames to=Language]{\transposedsingletaskcrossqatable}{\singletaskcrossqatable}
\pgfplotstabletypeset[columns/Language/.style={string type, column name={Language / Locale ID}},columns/Validation/.style={numeric type,precision=1,zerofill},columns/Test/.style={numeric type,precision=1,zerofill},
  every head row/.style={
    before row={\toprule},
    after row={\midrule}
  },
  every last row/.style={before row={\midrule}, after row=\bottomrule}
]{\transposedsingletaskcrossqatable}
\caption{Multiple-language compute-matched in-language QA results.}
\label{tab:multiple-lang-compute-matched-iqa}
\end{table}

\begin{table}
\pgfplotstableread{
Language ar	bn	fi	id	ko	ru	sw	te	en	Average
Validation 86.44940186	90.71659088	89.62218475	87.87414551	81.57859802	86.15505981	87.18105316	91.01767731	84.80833435	87.26700507
Test 87.74943542	87.6550293	89.66698456	88.70043182	87.48738098	88.74189758	90.52833557	92.92358398	87.90764618	89.0400806
}\singletaskcrossqatable
\pgfplotstabletranspose[string type, colnames from=Language, input colnames to=Language]{\transposedsingletaskcrossqatable}{\singletaskcrossqatable}
\pgfplotstabletypeset[columns/Language/.style={string type, column name={Language / Locale ID}},columns/Validation/.style={numeric type,precision=1,zerofill},columns/Test/.style={numeric type,precision=1,zerofill},
  every head row/.style={
    before row={\toprule},
    after row={\midrule}
  },
  every last row/.style={before row={\midrule}, after row=\bottomrule}
]{\transposedsingletaskcrossqatable}
\caption{Multiple-language parameter-matched in-language QA results.}
\label{tab:multiple-lang-param-matched-iqa}
\end{table}

\begin{table}
\pgfplotstableread{
Language ar	bn	fi	id	ko	ru	sw	te	en	Average
Validation 87.22318268	90.22686005	90.13412476	87.81632233	82.45137024	87.79477692	86.45631409	91.00112915	87.3582077	87.8291431
Test 89.10928345	89.12169647	89.72756958	88.8585968	85.90119934	89.11597443	90.21076202	93.52206421	88.86875153	89.38176643
}\singletaskcrossqatable
\pgfplotstabletranspose[string type, colnames from=Language, input colnames to=Language]{\transposedsingletaskcrossqatable}{\singletaskcrossqatable}
\pgfplotstabletypeset[columns/Language/.style={string type, column name={Language / Locale ID}},columns/Validation/.style={numeric type,precision=1,zerofill},columns/Test/.style={numeric type,precision=1,zerofill},
  every head row/.style={
    before row={\toprule},
    after row={\midrule}
  },
  every last row/.style={before row={\midrule}, after row=\bottomrule}
]{\transposedsingletaskcrossqatable}
\caption{\model in-language QA results.}
\label{tab:multiple-lang-flora-iqa}
\end{table}

\clearpage

\subsubsection{Semantic parsing}

\begin{table}
\pgfplotstableread{
Language am	be	bn	fi	ha	hu	ja	{pt-br}	ru	sw	ta	tr	yo	zu	{de-localized}	en	de	es	fr	hi	th	Average
Validation 25.52301216	34.72803497	38.49372482	33.89121246	28.87029266	30.96234322	36.82008362	36.82008362	40.58577347	34.72803497	31.38075256	40.16736221	20.50209236	25.10460281	31.18811798	37.66	33.89121246	35.26011658	30.10204124	35.48387146	36.47058868	33.26825497
Test 14.67964268	25.84482384	25.7366848	24.97972488	22.54663467	23.38469887	25.92592621	25.52041054	27.95350075	23.14138985	28.76453018	24.41200256	13.05758286	15.19329548	21.44772148	25.3	25.52041054	25.67675209	21.81749916	13.98809528	18.2567215	22.53085944
}\singletaskcrossqatable
\pgfplotstabletranspose[string type, colnames from=Language, input colnames to=Language]{\transposedsingletaskcrossqatable}{\singletaskcrossqatable}
\pgfplotstabletypeset[columns/Language/.style={string type, column name={Language / Locale ID}},columns/Validation/.style={numeric type,precision=1,zerofill},columns/Test/.style={numeric type,precision=1,zerofill},
  every head row/.style={
    before row={\toprule},
    after row={\midrule}
  },
  every last row/.style={before row={\midrule}, after row=\bottomrule}
]{\transposedsingletaskcrossqatable}
\caption{Single-language compute-matched semantic parsing results.}
\label{tab:single-lang-compute-matched-semparse}
\end{table}

\begin{table}
\pgfplotstableread{
Language am	be	bn	fi	ha	hu	ja	pt-br	ru	sw	ta	tr	yo	zu	de-localized	en	de	es	fr	hi	th	Average
Validation 41.00418472	52.30125427	51.04602432	53.13807678	46.02510452	50.62761688	52.30125427	57.74058533	60.66945648	51.04602432	47.28033447	53.13807678	39.33054352	41.00418472	57.4257431	54.81171417	54.81171417	57.8034668	62.75510025	54.83871078	55.29411697	52.11396608
Test 31.76534271	43.30900192	40.55150223	45.20140457	35.90159607	38.46985626	38.41578674	45.76912689	45.41767883	39.87564087	39.52419662	39.87564087	28.84563446	31.68423843	44.10187531	45.30954361	42.95755768	45.10350418	48.89222717	36.98979568	42.36773682	40.49185181
}\singletaskcrossqatable
\pgfplotstabletranspose[string type, colnames from=Language, input colnames to=Language]{\transposedsingletaskcrossqatable}{\singletaskcrossqatable}
\pgfplotstabletypeset[columns/Language/.style={string type, column name={Language / Locale ID}},columns/Validation/.style={numeric type,precision=1,zerofill},columns/Test/.style={numeric type,precision=1,zerofill},
  every head row/.style={
    before row={\toprule},
    after row={\midrule}
  },
  every last row/.style={before row={\midrule}, after row=\bottomrule}
]{\transposedsingletaskcrossqatable}
\caption{Multiple-language compute-matched semantic parsing results.}
\label{tab:multiple-lang-compute-matched-semparse}
\end{table}

\begin{table}
\pgfplotstableread{
Language am	be	bn	fi	ha	hu	ja	pt-br	ru	sw	ta	tr	yo	zu	de-localized	en	de	es	fr	hi	th	Average
Validation 48.53556442	58.99581528	60.66945648	62.34309769	53.55648422	60.25104523	56.06694412	61.92468643	64.43515015	51.88284683	57.32217407	59.41422653	43.51464462	44.76987457	65.84158325	64.01673889	58.15899658	63.00577927	65.81632996	65.80644989	65.29412079	58.64866711
Test 36.00973129	50.25682449	47.44525528	51.90592194	45.20140457	48.36442184	44.0929985	52.93322372	51.14896011	43.82265472	45.87726593	45.60692215	36.41524887	37.60475922	50.73726654	53.3387413	49.20248795	52.78662491	53.88659286	44.77040863	50.69384384	47.24293137
}\singletaskcrossqatable
\pgfplotstabletranspose[string type, colnames from=Language, input colnames to=Language]{\transposedsingletaskcrossqatable}{\singletaskcrossqatable}
\pgfplotstabletypeset[columns/Language/.style={string type, column name={Language / Locale ID}},columns/Validation/.style={numeric type,precision=1,zerofill},columns/Test/.style={numeric type,precision=1,zerofill},
  every head row/.style={
    before row={\toprule},
    after row={\midrule}
  },
  every last row/.style={before row={\midrule}, after row=\bottomrule}
]{\transposedsingletaskcrossqatable}
\caption{Multiple-language parameter-matched semantic parsing results.}
\label{tab:multiple-lang-param-matched-semparse}
\end{table}

\begin{table}
\pgfplotstableread{
Language am	be	bn	fi	ha	hu	ja	pt-br	ru	sw	ta	tr	yo	zu	de-localized	en	de	es	fr	hi	th	Average
Validation 47.69874573	58.57740402	54.39330673	59.41422653	49.37238312	56.48535538	56.90376663	60.66945648	60.66945648	58.57740402	55.64853668	58.15899658	44.35146332	37.23849487	61.88118744	64.85355377	54.39330673	61.27167511	62.75510025	62.58064651	59.41176605	56.44315393
Test 37.44255066	47.41822052	46.74236298	50.28385925	41.28142929	46.30981445	44.28223801	48.76993942	48.41849136	42.6872139	46.4720192	44.57961655	34.03622437	37.60475922	44.43699646	49.87834549	44.12003326	50.55732346	50.31918716	42.47449112	47.5281868	45.03063347
}\singletaskcrossqatable
\pgfplotstabletranspose[string type, colnames from=Language, input colnames to=Language]{\transposedsingletaskcrossqatable}{\singletaskcrossqatable}
\pgfplotstabletypeset[columns/Language/.style={string type, column name={Language / Locale ID}},columns/Validation/.style={numeric type,precision=1,zerofill},columns/Test/.style={numeric type,precision=1,zerofill},
  every head row/.style={
    before row={\toprule},
    after row={\midrule}
  },
  every last row/.style={before row={\midrule}, after row=\bottomrule}
]{\transposedsingletaskcrossqatable}
\caption{\model semantic parsing results.}
\label{tab:multiple-lang-flora-semparse}
\end{table}

\clearpage

\subsubsection{Named-entity recognition (NER)}

\begin{table}
\pgfplotstableread{
Language am	bm	bbj	ee	ha	ig	rw	lg	luo	mos	ny	pcm	sn	sw	tn	tw	wo	xh	yo	zu	Average
Validation 0.747692287	0.7337559462	0.6020260453	0.8491960168	0.8963460326	0.8195364475	0.778958261	0.8702594638	0.6613333225	0.6377584934	0.8680619001	0.821428597	0.8934487104	0.8960258961	0.7713534832	0.7290886641	0.8152793646	0.8373841047	0.7905632854	0.7916108966	0.7905553609
Test 0.7279151678	0.6889116764	0.6220703125	0.82405442	0.8649806976	0.8215174079	0.7609931231	0.8384409547	0.7144653797	0.6572360992	0.8830409646	0.8155540228	0.8864790201	0.8878628016	0.8437908292	0.7397683263	0.7647843361	0.8108866215	0.8071895242	0.8222523928	0.7891097039
}\singletaskcrossqatable
\pgfplotstabletranspose[string type, colnames from=Language, input colnames to=Language]{\transposedsingletaskcrossqatable}{\singletaskcrossqatable}
\pgfplotstabletypeset[columns/Language/.style={string type, column name={Language / Locale ID}},columns/Validation/.style={numeric type,precision=1,zerofill,preproc/expr={100*##1}},columns/Test/.style={numeric type,precision=1,zerofill,preproc/expr={100*##1}},
  every head row/.style={
    before row={\toprule},
    after row={\midrule}
  },
  every last row/.style={before row={\midrule}, after row=\bottomrule},
]{\transposedsingletaskcrossqatable}
\caption{Single-language compute-matched NER results.}
\label{tab:single-lang-compute-matched-ner}
\end{table}

\begin{table}
\pgfplotstableread{
Language am	bm	bbj	ee	ha	ig	rw	lg	luo	mos	ny	pcm	sn	sw	tn	tw	wo	xh	yo	zu	Average
Validation 0.7725191116	0.7236421704	0.5454545617	0.8289880753	0.9093632698	0.8435146213	0.7995813489	0.8748326898	0.7717391253	0.6757678986	0.8776579499	0.8284808397	0.9043775797	0.9066048861	0.8080807924	0.7670527697	0.8076256514	0.8348439336	0.8031188846	0.8466960192	0.8064971089
Test 0.7370303869	0.6776349545	0.6612658501	0.8255191445	0.8995633125	0.852611959	0.7738844156	0.8618322611	0.7629911304	0.6599212885	0.8919786215	0.8627042174	0.9015902281	0.9087711573	0.865701139	0.7912621498	0.7846846581	0.8378605247	0.8295739293	0.8732014298	0.8129791379
}\singletaskcrossqatable
\pgfplotstabletranspose[string type, colnames from=Language, input colnames to=Language]{\transposedsingletaskcrossqatable}{\singletaskcrossqatable}
\pgfplotstabletypeset[columns/Language/.style={string type, column name={Language / Locale ID}},columns/Validation/.style={numeric type,precision=1,zerofill,preproc/expr={100*##1}},columns/Test/.style={numeric type,precision=1,zerofill,preproc/expr={100*##1}},
  every head row/.style={
    before row={\toprule},
    after row={\midrule}
  },
  every last row/.style={before row={\midrule}, after row=\bottomrule},
]{\transposedsingletaskcrossqatable}
\caption{Multiple-language compute-matched NER results.}
\label{tab:multiple-lang-compute-matched-ner}
\end{table}

\begin{table}
\pgfplotstableread{
Language am	bm	bbj	ee	ha	ig	rw	lg	luo	mos	ny	pcm	sn	sw	tn	tw	wo	xh	yo	zu	Average
Validation 0.7914764285	0.7990353703	0.6574923396	0.8799291253	0.9358490705	0.8678929806	0.8213914633	0.8918188214	0.767123282	0.720757246	0.8877836466	0.8738251328	0.9269496799	0.9174947739	0.8290086389	0.8149134517	0.8110281229	0.8539553881	0.8223844171	0.8571428657	0.8363626122
Test 0.7736185193	0.7384295464	0.6828528047	0.8666089177	0.9125874043	0.8516329527	0.7795585394	0.8627333045	0.7708333135	0.7055555582	0.8917539716	0.882291317	0.9261060953	0.9078138471	0.8856672049	0.817589581	0.8084722757	0.861669004	0.8469302058	0.872908175	0.8322806269
}\singletaskcrossqatable
\pgfplotstabletranspose[string type, colnames from=Language, input colnames to=Language]{\transposedsingletaskcrossqatable}{\singletaskcrossqatable}
\pgfplotstabletypeset[columns/Language/.style={string type, column name={Language / Locale ID}},columns/Validation/.style={numeric type,precision=1,zerofill,preproc/expr={100*##1}},columns/Test/.style={numeric type,precision=1,zerofill,preproc/expr={100*##1}},
  every head row/.style={
    before row={\toprule},
    after row={\midrule}
  },
  every last row/.style={before row={\midrule}, after row=\bottomrule},
]{\transposedsingletaskcrossqatable}
\caption{Multiple-language parameter-matched NER results.}
\label{tab:multiple-lang-param-matched-ner}
\end{table}

\begin{table}
\pgfplotstableread{
Language am	bm	bbj	ee	ha	ig	rw	lg	luo	mos	ny	pcm	sn	sw	tn	tw	wo	xh	yo	zu	Average
Validation 0.8181818128	0.7924528122	0.6647058725	0.8717504144	0.9364248514	0.8594142199	0.8347381949	0.9023501873	0.7862796783	0.7298887968	0.8950579762	0.8715519905	0.9319242835	0.915035665	0.830530405	0.8118556738	0.8472850919	0.8673316836	0.8345187306	0.8653931618	0.8433335751
Test 0.8000000119	0.7316326499	0.7209185958	0.860904634	0.9108225107	0.8646986485	0.8004715443	0.8807556033	0.8035044074	0.7365010977	0.9182323217	0.8787982464	0.9318650961	0.9106791615	0.8809061646	0.7958366871	0.8128628731	0.8704835773	0.8602762222	0.8904719353	0.8430310994
}\singletaskcrossqatable
\pgfplotstabletranspose[string type, colnames from=Language, input colnames to=Language]{\transposedsingletaskcrossqatable}{\singletaskcrossqatable}
\pgfplotstabletypeset[columns/Language/.style={string type, column name={Language / Locale ID}},columns/Validation/.style={numeric type,precision=1,zerofill,preproc/expr={100*##1}},columns/Test/.style={numeric type,precision=1,zerofill,preproc/expr={100*##1}},
  every head row/.style={
    before row={\toprule},
    after row={\midrule}
  },
  every last row/.style={before row={\midrule}, after row=\bottomrule},
]{\transposedsingletaskcrossqatable}
\caption{\model NER results.}
\label{tab:multiple-lang-flora-ner}
\end{table}

\clearpage

\subsection{Multitask experiments}

\begin{table}
\pgfplotstableread{
Task SemParse	QA-InLang	NER	Average
Validation 34.72803497	84.1835022	88.63530755	69.18228157
Test 22.95214844	87.32514191	88.74230385	66.33986473
}\singletaskcrossqatable
\pgfplotstabletranspose[string type, colnames from=Task, input colnames to=Task]{\transposedsingletaskcrossqatable}{\singletaskcrossqatable}
\pgfplotstabletypeset[columns/Task/.style={string type, column name={Task}},columns/Validation/.style={numeric type,precision=1,zerofill,preproc/expr={1*##1}},columns/Test/.style={numeric type,precision=1,zerofill,preproc/expr={1*##1}},
  every head row/.style={
    before row={\toprule},
    after row={\midrule}
  },
  every last row/.style={before row={\midrule}, after row=\bottomrule},
]{\transposedsingletaskcrossqatable}
\caption{Single-task compute-matched Swahili results.}
\label{tab:single-lang-compute-matched-sw}
\end{table}

\begin{table}
\pgfplotstableread{
Task SemParse	QA-InLang	NER	Average
Validation 37.65690231	83.945755	88.63741159	70.08002297
Test 23.03325272	86.42139435	88.67841363	66.04435356
}\singletaskcrossqatable
\pgfplotstabletranspose[string type, colnames from=Task, input colnames to=Task]{\transposedsingletaskcrossqatable}{\singletaskcrossqatable}
\pgfplotstabletypeset[columns/Task/.style={string type, column name={Task}},columns/Validation/.style={numeric type,precision=1,zerofill,preproc/expr={1*##1}},columns/Test/.style={numeric type,precision=1,zerofill,preproc/expr={1*##1}},
  every head row/.style={
    before row={\toprule},
    after row={\midrule}
  },
  every last row/.style={before row={\midrule}, after row=\bottomrule},
]{\transposedsingletaskcrossqatable}
\caption{Multiple-task compute-matched Swahili results.}
\label{tab:multiple-task-compute-matched-sw}
\end{table}

\begin{table}
\pgfplotstableread{
Task SemParse	QA-InLang	NER	Average
Validation 38.49372482	82.54894257	89.72948194	70.25738311
Test 26.41254425	85.1645813	88.75687718	66.77800091
}\singletaskcrossqatable
\pgfplotstabletranspose[string type, colnames from=Task, input colnames to=Task]{\transposedsingletaskcrossqatable}{\singletaskcrossqatable}
\pgfplotstabletypeset[columns/Task/.style={string type, column name={Task}},columns/Validation/.style={numeric type,precision=1,zerofill,preproc/expr={1*##1}},columns/Test/.style={numeric type,precision=1,zerofill,preproc/expr={1*##1}},
  every head row/.style={
    before row={\toprule},
    after row={\midrule}
  },
  every last row/.style={before row={\midrule}, after row=\bottomrule},
]{\transposedsingletaskcrossqatable}
\caption{Multiple-task parameter-matched Swahili results.}
\label{tab:multiple-task-param-matched-sw}
\end{table}

\begin{table}
\pgfplotstableread{
Task SemParse	QA-InLang	NER	Average
Validation 47.69874573	85.20032501	90.96744657	74.62217244
Test 32.65747452	87.47990417	90.31975269	70.15237713
}\singletaskcrossqatable
\pgfplotstabletranspose[string type, colnames from=Task, input colnames to=Task]{\transposedsingletaskcrossqatable}{\singletaskcrossqatable}
\pgfplotstabletypeset[columns/Task/.style={string type, column name={Task}},columns/Validation/.style={numeric type,precision=1,zerofill,preproc/expr={1*##1}},columns/Test/.style={numeric type,precision=1,zerofill,preproc/expr={1*##1}},
  every head row/.style={
    before row={\toprule},
    after row={\midrule}
  },
  every last row/.style={before row={\midrule}, after row=\bottomrule},
]{\transposedsingletaskcrossqatable}
\caption{\model Swahili results.}
\label{tab:multiple-task-flora-sw}
\end{table}

\begin{table}
\pgfplotstableread{
Task SemParse	QA-InLang	QA-CrossLang	Average
Validation 37.23849487	87.62257385	78.59348297	67.8181839
Test 25.7366848	86.6550293	83.22580719	65.20584043
}\singletaskcrossqatable
\pgfplotstabletranspose[string type, colnames from=Task, input colnames to=Task]{\transposedsingletaskcrossqatable}{\singletaskcrossqatable}
\pgfplotstabletypeset[columns/Task/.style={string type, column name={Task}},columns/Validation/.style={numeric type,precision=1,zerofill,preproc/expr={1*##1}},columns/Test/.style={numeric type,precision=1,zerofill,preproc/expr={1*##1}},
  every head row/.style={
    before row={\toprule},
    after row={\midrule}
  },
  every last row/.style={before row={\midrule}, after row=\bottomrule},
]{\transposedsingletaskcrossqatable}
\caption{Single-task compute-matched Bengali results.}
\label{tab:single-lang-compute-matched-bn}
\end{table}

\begin{table}
\pgfplotstableread{
Task SemParse	QA-InLang	QA-CrossLang	Average
Validation 36.40167236	88.44062042	76.63960266	67.16063182
Test 24.27683067	85.36190796	81.87148285	63.83674049
}\singletaskcrossqatable
\pgfplotstabletranspose[string type, colnames from=Task, input colnames to=Task]{\transposedsingletaskcrossqatable}{\singletaskcrossqatable}
\pgfplotstabletypeset[columns/Task/.style={string type, column name={Task}},columns/Validation/.style={numeric type,precision=1,zerofill,preproc/expr={1*##1}},columns/Test/.style={numeric type,precision=1,zerofill,preproc/expr={1*##1}},
  every head row/.style={
    before row={\toprule},
    after row={\midrule}
  },
  every last row/.style={before row={\midrule}, after row=\bottomrule},
]{\transposedsingletaskcrossqatable}
\caption{Multiple-task compute-matched Bengali results.}
\label{tab:multiple-task-compute-matched-bn}
\end{table}

\begin{table}
\pgfplotstableread{
Task SemParse	QA-InLang	QA-CrossLang	Average
Validation 44.35146332	89.06707764	76.07772064	69.8320872
Test 29.35928535	83.64680481	84.35270691	65.78626569
}\singletaskcrossqatable
\pgfplotstabletranspose[string type, colnames from=Task, input colnames to=Task]{\transposedsingletaskcrossqatable}{\singletaskcrossqatable}
\pgfplotstabletypeset[columns/Task/.style={string type, column name={Task}},columns/Validation/.style={numeric type,precision=1,zerofill,preproc/expr={1*##1}},columns/Test/.style={numeric type,precision=1,zerofill,preproc/expr={1*##1}},
  every head row/.style={
    before row={\toprule},
    after row={\midrule}
  },
  every last row/.style={before row={\midrule}, after row=\bottomrule},
]{\transposedsingletaskcrossqatable}
\caption{Multiple-task parameter-matched Bengali results.}
\label{tab:multiple-task-param-matched-bn}
\end{table}

\begin{table}
\pgfplotstableread{
Task SemParse	QA-InLang	QA-CrossLang	Average
Validation 52.71966553	89.99461365	77.2778244	73.33070119
Test 37.01000214	85.87830353	84.07333374	68.98721313
}\singletaskcrossqatable
\pgfplotstabletranspose[string type, colnames from=Task, input colnames to=Task]{\transposedsingletaskcrossqatable}{\singletaskcrossqatable}
\pgfplotstabletypeset[columns/Task/.style={string type, column name={Task}},columns/Validation/.style={numeric type,precision=1,zerofill,preproc/expr={1*##1}},columns/Test/.style={numeric type,precision=1,zerofill,preproc/expr={1*##1}},
  every head row/.style={
    before row={\toprule},
    after row={\midrule}
  },
  every last row/.style={before row={\midrule}, after row=\bottomrule},
]{\transposedsingletaskcrossqatable}
\caption{\model Bengali results.}
\label{tab:multiple-task-flora-bn}
\end{table}

\clearpage

\section{Full results of \cref{tab:multilingual-multitask-results}}
\label{sec:joint-mtml-full}

As in \cref{sec:single-feature-full}, we include validation results along with test results.

\subsection{Supervised learning results}

\subsubsection{Cross-lingual QA}

\begin{table}
\pgfplotstableread{
Language ar	bn	fi	ko	ru	te	Average
Validation 83.63832855	79.60105133	81.36396027	84.96627045	79.79834747	80.32051086	81.61474482
Test 82.2605896	82.9102478	82.21983337	85.24206543	81.67617798	84.37302399	83.11365636
}\singletaskcrossqatable
\pgfplotstabletranspose[string type, colnames from=Language, input colnames to=Language]{\transposedsingletaskcrossqatable}{\singletaskcrossqatable}
\pgfplotstabletypeset[columns/Language/.style={string type, column name={Language / Locale ID}},columns/Validation/.style={numeric type,precision=1,zerofill},columns/Test/.style={numeric type,precision=1,zerofill},
  every head row/.style={
    before row={\toprule},
    after row={\midrule}
  },
  every last row/.style={before row={\midrule}, after row=\bottomrule}
]{\transposedsingletaskcrossqatable}
\caption{Multitask multilingual compute-matched cross-lingual QA results.}
\label{tab:mtml-compute-matched-cqa}
\end{table}

\begin{table}
\pgfplotstableread{
Language ar	bn	fi	ko	ru	te	Average
Validation 82.11297607	79.26996613	81.76979828	84.51812744	79.46916962	81.03040314	81.36174011
Test 83.60165405	84.51586914	81.33002472	86.04283905	82.6519165	80.99311829	83.18923696
}\singletaskcrossqatable
\pgfplotstabletranspose[string type, colnames from=Language, input colnames to=Language]{\transposedsingletaskcrossqatable}{\singletaskcrossqatable}
\pgfplotstabletypeset[columns/Language/.style={string type, column name={Language / Locale ID}},columns/Validation/.style={numeric type,precision=1,zerofill},columns/Test/.style={numeric type,precision=1,zerofill},
  every head row/.style={
    before row={\toprule},
    after row={\midrule}
  },
  every last row/.style={before row={\midrule}, after row=\bottomrule}
]{\transposedsingletaskcrossqatable}
\caption{Multitask multilingual  parameter-matched cross-lingual QA results.}
\label{tab:mtml-param-matched-cqa}
\end{table}

\begin{table}
\pgfplotstableread{
Language ar	bn	fi	ko	ru	te	Average
Validation 85.56196594	82.19650269	82.59798431	85.54367828	80.51891327	82.11548615	83.08908844
Test 84.32571411	86.24279785	83.17665863	88.22769928	84.04055023	85.02754211	85.1734937
}\singletaskcrossqatable
\pgfplotstabletranspose[string type, colnames from=Language, input colnames to=Language]{\transposedsingletaskcrossqatable}{\singletaskcrossqatable}
\pgfplotstabletypeset[columns/Language/.style={string type, column name={Language / Locale ID}},columns/Validation/.style={numeric type,precision=1,zerofill},columns/Test/.style={numeric type,precision=1,zerofill},
  every head row/.style={
    before row={\toprule},
    after row={\midrule}
  },
  every last row/.style={before row={\midrule}, after row=\bottomrule}
]{\transposedsingletaskcrossqatable}
\caption{Multitask multilingual \model cross-lingual QA results.}
\label{tab:mtml-flora-cqa}
\end{table}

\clearpage
\subsubsection{In-language QA}

\begin{table}
\pgfplotstableread{
Language ar	bn	fi	id	ko	ru	sw	te	en	Average
Validation 83.77384186	88.10728455	86.70277405	86.53463745	80.9304657	82.63617706	85.17017365	89.16565704	83.32946014	85.15005239
Test 84.88231659	83.85608673	87.38994598	88.14199829	85.0464859	84.07277679	88.2199173	91.66433716	85.85596466	86.56998105
}\singletaskcrossqatable
\pgfplotstabletranspose[string type, colnames from=Language, input colnames to=Language]{\transposedsingletaskcrossqatable}{\singletaskcrossqatable}
\pgfplotstabletypeset[columns/Language/.style={string type, column name={Language / Locale ID}},columns/Validation/.style={numeric type,precision=1,zerofill},columns/Test/.style={numeric type,precision=1,zerofill},
  every head row/.style={
    before row={\toprule},
    after row={\midrule}
  },
  every last row/.style={before row={\midrule}, after row=\bottomrule}
]{\transposedsingletaskcrossqatable}
\caption{Multitask multilingual  compute-matched in-language QA results.}
\label{tab:mtml-compute-matched-iqa}
\end{table}

\begin{table}
\pgfplotstableread{
Language ar	bn	fi	id	ko	ru	sw	te	en	Average
Validation 85.1965332	90.82045746	86.99008179	85.37218475	80.35877228	84.86812592	84.75176239	89.40311432	84.53672028	85.81086138
Test 86.92311096	85.80987549	87.46871185	87.14300537	85.32518005	86.22168732	88.70110321	91.72825623	86.66725922	87.33202108
}\singletaskcrossqatable
\pgfplotstabletranspose[string type, colnames from=Language, input colnames to=Language]{\transposedsingletaskcrossqatable}{\singletaskcrossqatable}
\pgfplotstabletypeset[columns/Language/.style={string type, column name={Language / Locale ID}},columns/Validation/.style={numeric type,precision=1,zerofill},columns/Test/.style={numeric type,precision=1,zerofill},
  every head row/.style={
    before row={\toprule},
    after row={\midrule}
  },
  every last row/.style={before row={\midrule}, after row=\bottomrule}
]{\transposedsingletaskcrossqatable}
\caption{Multitask multilingual  parameter-matched in-language QA results.}
\label{tab:mtml-param-matched-iqa}
\end{table}

\begin{table}
\pgfplotstableread{
Language ar	bn	fi	id	ko	ru	sw	te	en	Average
Validation 86.36582184	90.36032104	90.00484467	88.64551544	83.19483185	85.39755249	87.35567474	91.54283905	84.99828339	87.54063161
Test 88.71115112	91.06983948	89.52046204	89.4889679	86.09911346	88.49821472	90.11864471	93.55233002	87.92835236	89.44300842
}\singletaskcrossqatable
\pgfplotstabletranspose[string type, colnames from=Language, input colnames to=Language]{\transposedsingletaskcrossqatable}{\singletaskcrossqatable}
\pgfplotstabletypeset[columns/Language/.style={string type, column name={Language / Locale ID}},columns/Validation/.style={numeric type,precision=1,zerofill},columns/Test/.style={numeric type,precision=1,zerofill},
  every head row/.style={
    before row={\toprule},
    after row={\midrule}
  },
  every last row/.style={before row={\midrule}, after row=\bottomrule}
]{\transposedsingletaskcrossqatable}
\caption{Multitask multilingual \model in-language QA results.}
\label{tab:mtml-flora-iqa}
\end{table}

\clearpage
\subsubsection{Semantic parsing}

\begin{table}
\pgfplotstableread{
Language am	be	bn	fi	ha	hu	ja	pt-br	ru	sw	ta	tr	yo	zu	de-localized	en	de	es	fr	hi	th	Average
Validation 35.98326492	47.69874573	51.46443558	53.55648422	39.33054352	50.20920563	46.44351578	53.55648422	53.97489548	46.44351578	43.51464462	45.18828583	29.70711327	33.05439377	56.43564224	50.62761688	47.69874573	54.91329575	55.61224365	49.67741776	54.70588303	47.60935111
Test 24.54717445	38.5779953	34.27953339	40.06488419	29.98107529	36.87483215	33.90105438	41.06515121	40.22708893	34.33360291	31.3868618	33.65774536	23.97945404	29.27818298	40.31501389	39.90267563	37.98323822	42.19745255	41.94517517	31.37755013	37.46747589	35.39729609
}\singletaskcrossqatable
\pgfplotstabletranspose[string type, colnames from=Language, input colnames to=Language]{\transposedsingletaskcrossqatable}{\singletaskcrossqatable}
\pgfplotstabletypeset[columns/Language/.style={string type, column name={Language / Locale ID}},columns/Validation/.style={numeric type,precision=1,zerofill},columns/Test/.style={numeric type,precision=1,zerofill},
  every head row/.style={
    before row={\toprule},
    after row={\midrule}
  },
  every last row/.style={before row={\midrule}, after row=\bottomrule}
]{\transposedsingletaskcrossqatable}
\caption{Multitask multilingual  compute-matched semantic parsing results.}
\label{tab:mtml-compute-matched-semparse}
\end{table}

\begin{table}
\pgfplotstableread{
Language am	be	bn	fi	ha	hu	ja	pt-br	ru	sw	ta	tr	yo	zu	de-localized	en	de	es	fr	hi	th	Average
Validation 43.09623337	58.57740402	58.99581528	62.34309769	51.04602432	58.15899658	58.99581528	64.01673889	63.17991638	55.64853668	53.97489548	58.57740402	43.51464462	46.86192322	61.88118744	61.50627518	56.06694412	64.73988342	62.75510025	66.45161438	63.52941132	57.80561247
Test 35.52311325	50.41903305	47.39118576	52.96025848	42.11949158	47.22898102	45.87726593	51.71667862	51.33819962	44.06596375	47.79670334	46.58015823	34.25250244	39.71343613	49.76541519	53.12246704	48.55366135	51.71178436	53.69883728	46.42856979	49.65307999	47.13889458
}\singletaskcrossqatable
\pgfplotstabletranspose[string type, colnames from=Language, input colnames to=Language]{\transposedsingletaskcrossqatable}{\singletaskcrossqatable}
\pgfplotstabletypeset[columns/Language/.style={string type, column name={Language / Locale ID}},columns/Validation/.style={numeric type,precision=1,zerofill},columns/Test/.style={numeric type,precision=1,zerofill},
  every head row/.style={
    before row={\toprule},
    after row={\midrule}
  },
  every last row/.style={before row={\midrule}, after row=\bottomrule}
]{\transposedsingletaskcrossqatable}
\caption{Multitask multilingual  parameter-matched semantic parsing results.}
\label{tab:mtml-param-matched-semparse}
\end{table}

\begin{table}
\pgfplotstableread{
Language am	be	bn	fi	ha	hu	ja	pt-br	ru	sw	ta	tr	yo	zu	de-localized	en	de	es	fr	hi	th	Average
Validation 46.86192322	59.83263779	58.15899658	59.83263779	50.62761688	55.23012543	53.97489548	61.50627518	62.76150513	55.23012543	53.13807678	57.74058533	43.51464462	41.42259598	60.39603806	61.08786774	57.74058533	60.69364166	65.81632996	63.87096786	61.76470566	56.7239418
Test 38.253582	48.82400513	47.12084198	50.87861633	41.03812027	44.17410278	45.47174835	47.58042526	50.60827255	42.57907486	45.66098785	45.44471359	34.09029388	37.57772446	45.64343262	50.68937683	45.49878311	50.39809036	52.49718475	45.19557953	48.17866516	45.59064865
}\singletaskcrossqatable
\pgfplotstabletranspose[string type, colnames from=Language, input colnames to=Language]{\transposedsingletaskcrossqatable}{\singletaskcrossqatable}
\pgfplotstabletypeset[columns/Language/.style={string type, column name={Language / Locale ID}},columns/Validation/.style={numeric type,precision=1,zerofill},columns/Test/.style={numeric type,precision=1,zerofill},
  every head row/.style={
    before row={\toprule},
    after row={\midrule}
  },
  every last row/.style={before row={\midrule}, after row=\bottomrule}
]{\transposedsingletaskcrossqatable}
\caption{Multitask multilingual \model semantic parsing results.}
\label{tab:mtml-flora-semparse}
\end{table}

\clearpage
\subsubsection{NER}

\begin{table}
\pgfplotstableread{
Language am	bm	bbj	ee	ha	ig	rw	lg	luo	mos	ny	pcm	sn	sw	tn	tw	wo	xh	yo	zu	Average
Validation 0.6883720756	0.6726591587	0.5244866014	0.8052299619	0.8746177554	0.7823157907	0.7720588446	0.854257822	0.6170799136	0.6287625432	0.8343104124	0.7992070317	0.841226995	0.8720795512	0.7777048945	0.7404580116	0.751614809	0.7329414487	0.7144268751	0.7540106773	0.7518910587
Test 0.7081081271	0.634577632	0.6119710803	0.7858384252	0.8481124043	0.821328342	0.7158403993	0.8237063885	0.6692606807	0.6156424284	0.8551427126	0.8128231764	0.8436632752	0.8707333207	0.8430989385	0.7476038337	0.7394429445	0.7716136575	0.7501035929	0.7872244716	0.7627917916
}\singletaskcrossqatable
\pgfplotstabletranspose[string type, colnames from=Language, input colnames to=Language]{\transposedsingletaskcrossqatable}{\singletaskcrossqatable}
\pgfplotstabletypeset[columns/Language/.style={string type, column name={Language / Locale ID}},columns/Validation/.style={numeric type,precision=1,zerofill,preproc/expr={100*##1}},columns/Test/.style={numeric type,precision=1,zerofill,preproc/expr={100*##1}},
  every head row/.style={
    before row={\toprule},
    after row={\midrule}
  },
  every last row/.style={before row={\midrule}, after row=\bottomrule},
]{\transposedsingletaskcrossqatable}
\caption{Multitask multilingual compute-matched NER results.}
\label{tab:mtml-compute-matched-ner}
\end{table}

\begin{table}
\pgfplotstableread{
Language am	bm	bbj	ee	ha	ig	rw	lg	luo	mos	ny	pcm	sn	sw	tn	tw	wo	xh	yo	zu	Average
Validation 0.7442572713	0.7507553101	0.6174699068	0.8506151438	0.9210925698	0.8540084362	0.8179927468	0.8734006882	0.770889461	0.6671131849	0.871463716	0.854379952	0.9180005193	0.9068033695	0.8204455972	0.7836411595	0.7944412231	0.8244236112	0.7947342992	0.8288770318	0.8132402599
Test 0.7671480179	0.710252583	0.6612740159	0.8346286416	0.9054113626	0.8611369729	0.7676585317	0.8533812165	0.7589743733	0.6637313962	0.8787797689	0.8691099286	0.9061922431	0.9014582634	0.8764705658	0.7788617611	0.7894265056	0.8553776741	0.8121363521	0.8608852625	0.8156147718
}\singletaskcrossqatable
\pgfplotstabletranspose[string type, colnames from=Language, input colnames to=Language]{\transposedsingletaskcrossqatable}{\singletaskcrossqatable}
\pgfplotstabletypeset[columns/Language/.style={string type, column name={Language / Locale ID}},columns/Validation/.style={numeric type,precision=1,zerofill,preproc/expr={100*##1}},columns/Test/.style={numeric type,precision=1,zerofill,preproc/expr={100*##1}},
  every head row/.style={
    before row={\toprule},
    after row={\midrule}
  },
  every last row/.style={before row={\midrule}, after row=\bottomrule},
]{\transposedsingletaskcrossqatable}
\caption{Multitask multilingual parameter-matched NER results.}
\label{tab:mtml-param-matched-ner}
\end{table}

\begin{table}
\pgfplotstableread{
Language am	bm	bbj	ee	ha	ig	rw	lg	luo	mos	ny	pcm	sn	sw	tn	tw	wo	xh	yo	zu	Average
Validation 0.8111454844	0.8044871688	0.6360946894	0.870370388	0.9270128012	0.8536378741	0.8389917612	0.8909029961	0.7474226952	0.7117056847	0.8936959505	0.8677130342	0.9186791778	0.9164345264	0.8318703771	0.7943444848	0.8326624632	0.8554247022	0.8298391104	0.848538518	0.8340486944
Test 0.7938237786	0.7372708917	0.7048282623	0.8607813716	0.9119442105	0.8685358167	0.8021680117	0.8718102574	0.7888748646	0.7253481746	0.9040214419	0.8891976476	0.9273356199	0.9144243598	0.8885941505	0.7824701071	0.8187556267	0.8603414297	0.8584551215	0.8866906762	0.839783591
}\singletaskcrossqatable
\pgfplotstabletranspose[string type, colnames from=Language, input colnames to=Language]{\transposedsingletaskcrossqatable}{\singletaskcrossqatable}
\pgfplotstabletypeset[columns/Language/.style={string type, column name={Language / Locale ID}},columns/Validation/.style={numeric type,precision=1,zerofill,preproc/expr={100*##1}},columns/Test/.style={numeric type,precision=1,zerofill,preproc/expr={100*##1}},
  every head row/.style={
    before row={\toprule},
    after row={\midrule}
  },
  every last row/.style={before row={\midrule}, after row=\bottomrule},
]{\transposedsingletaskcrossqatable}
\caption{Multitask multilingual \model NER results.}
\label{tab:mtml-flora-ner}
\end{table}

\clearpage

\subsection{Zero-shot results}

\subsubsection{Zero-shot unseen combinations: cross-lingual QA}

\begin{table}
\pgfplotstableread{
Language hi ta Average
Validation 82.0928421	79.27688599 80.68486404
Test 83.21633911	80.07525635 81.64579773
}\singletaskcrossqatable
\pgfplotstabletranspose[string type, colnames from=Language, input colnames to=Language]{\transposedsingletaskcrossqatable}{\singletaskcrossqatable}
\pgfplotstabletypeset[columns/Language/.style={string type, column name={Language / Locale ID}},columns/Validation/.style={numeric type,precision=1,zerofill},columns/Test/.style={numeric type,precision=1,zerofill},
  every head row/.style={
    before row={\toprule},
    after row={\midrule}
  },
  every last row/.style={before row={\midrule}, after row=\bottomrule}
]{\transposedsingletaskcrossqatable}
\caption{Zero-shot unseen combinations: compute-matched cross-lingual QA results.}
\label{tab:zeroshot-unseen-comb-compute-matched-cqa}
\end{table}

\begin{table}
\pgfplotstableread{
Language hi ta Average
Validation 84.35079193	80.65543365	82.50311279
Test 84.38761902	79.84228516	82.11495209
}\singletaskcrossqatable
\pgfplotstabletranspose[string type, colnames from=Language, input colnames to=Language]{\transposedsingletaskcrossqatable}{\singletaskcrossqatable}
\pgfplotstabletypeset[columns/Language/.style={string type, column name={Language / Locale ID}},columns/Validation/.style={numeric type,precision=1,zerofill},columns/Test/.style={numeric type,precision=1,zerofill},
  every head row/.style={
    before row={\toprule},
    after row={\midrule}
  },
  every last row/.style={before row={\midrule}, after row=\bottomrule}
]{\transposedsingletaskcrossqatable}
\caption{Zero-shot unseen combinations: parameter-matched cross-lingual QA results.}
\label{tab:zeroshot-unseen-comb-param-matched-cqa}
\end{table}

\begin{table}
\pgfplotstableread{
Language hi ta Average
Validation 84.93553162	81.93753052 83.43653107
Test 85.2538147	80.41938782 82.83660126
}\singletaskcrossqatable
\pgfplotstabletranspose[string type, colnames from=Language, input colnames to=Language]{\transposedsingletaskcrossqatable}{\singletaskcrossqatable}
\pgfplotstabletypeset[columns/Language/.style={string type, column name={Language / Locale ID}},columns/Validation/.style={numeric type,precision=1,zerofill},columns/Test/.style={numeric type,precision=1,zerofill},
  every head row/.style={
    before row={\toprule},
    after row={\midrule}
  },
  every last row/.style={before row={\midrule}, after row=\bottomrule}
]{\transposedsingletaskcrossqatable}
\caption{Zero-shot unseen combinations: \model cross-lingual QA results.}
\label{tab:zeroshot-unseen-comb-flora-cqa}
\end{table}

\clearpage

\subsubsection{Zero-shot unseen combinations: semantic parsing}

\begin{table}
\pgfplotstableread{
Language am	be	bn	ha	sw	ta	yo	zu	th	Average
Validation 13.80753136	35.56485367	30.96234322	20.08368111	34.30962372	20.50209236	7.531380653	17.15481186	37.64706039	24.1737087
Test 10.19194412	30.14328194	21.57339859	16.16653061	21.97891235	11.40848923	8.948364258	15.97729111	22.98352051	17.7079703
}\singletaskcrossqatable
\pgfplotstabletranspose[string type, colnames from=Language, input colnames to=Language]{\transposedsingletaskcrossqatable}{\singletaskcrossqatable}
\pgfplotstabletypeset[columns/Language/.style={string type, column name={Language / Locale ID}},columns/Validation/.style={numeric type,precision=1,zerofill},columns/Test/.style={numeric type,precision=1,zerofill},
  every head row/.style={
    before row={\toprule},
    after row={\midrule}
  },
  every last row/.style={before row={\midrule}, after row=\bottomrule}
]{\transposedsingletaskcrossqatable}
\caption{Zero-shot unseen combinations: compute-matched semantic parsing results.}
\label{tab:zeroshot-unseen-comb-compute-matched-semparse}
\end{table}

\begin{table}
\pgfplotstableread{
Language am	be	bn	ha	sw	ta	yo	zu	th	average
Validation 24.68619156	53.97489548	45.18828583	30.54393387	43.09623337	30.54393387	21.33891296	27.61506271	55.88235474	36.98553382
Test 17.57231712	42.76831436	35.55014801	25.46634293	32.17085648	23.06028748	16.13949776	24.81751823	38.37814331	28.43593619
}\singletaskcrossqatable
\pgfplotstabletranspose[string type, colnames from=Language, input colnames to=Language]{\transposedsingletaskcrossqatable}{\singletaskcrossqatable}
\pgfplotstabletypeset[columns/Language/.style={string type, column name={Language / Locale ID}},columns/Validation/.style={numeric type,precision=1,zerofill},columns/Test/.style={numeric type,precision=1,zerofill},
  every head row/.style={
    before row={\toprule},
    after row={\midrule}
  },
  every last row/.style={before row={\midrule}, after row=\bottomrule}
]{\transposedsingletaskcrossqatable}
\caption{Zero-shot unseen combinations: parameter-matched semantic parsing results.}
\label{tab:zeroshot-unseen-comb-param-matched-semparse}
\end{table}

\begin{table}
\pgfplotstableread{
Language am	be	bn	ha	sw	ta	yo	zu	th	Average
Validation 46.86192322	59.83263779	58.15899658	50.62761688	55.23012543	53.13807678	43.51464462	41.42259598	61.76470566	52.28348033
Test 38.253582	48.82400513	47.12084198	41.03812027	42.57907486	45.66098785	34.09029388	37.57772446	48.17866516	42.59147729
}\singletaskcrossqatable
\pgfplotstabletranspose[string type, colnames from=Language, input colnames to=Language]{\transposedsingletaskcrossqatable}{\singletaskcrossqatable}
\pgfplotstabletypeset[columns/Language/.style={string type, column name={Language / Locale ID}},columns/Validation/.style={numeric type,precision=1,zerofill},columns/Test/.style={numeric type,precision=1,zerofill},
  every head row/.style={
    before row={\toprule},
    after row={\midrule}
  },
  every last row/.style={before row={\midrule}, after row=\bottomrule}
]{\transposedsingletaskcrossqatable}
\caption{Zero-shot unseen combinations: \model semantic parsing results.}
\label{tab:zeroshot-unseen-comb-flora-semparse}
\end{table}

\clearpage

\subsubsection{Zero-shot unseen languages}

\begin{table}
\pgfplotstableread{
Language as	bho	brx	gbm	gom	gu	hne	kn	mai	ml	mni	mr	mwr	or	pa	ps	sa	ur	Average
Validation 80.73505402	78.92473602	45.74462891	74.64048004	77.75396729	80.8904953	77.05461121	80.89189148	79.30569458	80.97523499	52.29472733	78.70146942	80.35784912	77.77970886	79.3065033	76.97264862	78.14186859	79.25789642	75.54052586
Test 79.61084747	76.02685547	40.95887756	73.11741638	75.25467682	79.80988312	78.48851776	80.46699524	76.21376801	81.36724854	52.91281128	78.65423584	76.88059998	77.47741699	80.38660431	78.24221039	80.48957062	78.53969574	74.71656842
}\singletaskcrossqatable
\pgfplotstabletranspose[string type, colnames from=Language, input colnames to=Language]{\transposedsingletaskcrossqatable}{\singletaskcrossqatable}
\pgfplotstabletypeset[columns/Language/.style={string type, column name={Language / Locale ID}},columns/Validation/.style={numeric type,precision=1,zerofill},columns/Test/.style={numeric type,precision=1,zerofill},
  every head row/.style={
    before row={\toprule},
    after row={\midrule}
  },
  every last row/.style={before row={\midrule}, after row=\bottomrule}
]{\transposedsingletaskcrossqatable}
\caption{Zero-shot unseen languages: compute-matched cross-lingual QA results.}
\label{tab:zeroshot-unseen-lang-compute-matched-cqa}
\end{table}

\begin{table}
\pgfplotstableread{
Language as	bho	brx	gbm	gom	gu	hne	kn	mai	ml	mni	mr	mwr	or	pa	ps	sa	ur	Average
Validation 81.28739166	79.02025604	46.45176697	73.35533905	77.71311951	81.85476685	78.1129837	82.67718506	81.31371307	81.61295319	48.82740021	79.96716309	77.73997498	77.83130646	80.03722382	77.04194641	77.49008942	78.32826233	75.5923801
Test 79.78107452	77.02669525	41.18789291	72.46368408	74.72678375	80.00364685	77.22098541	80.07883453	77.50582123	80.93243408	51.09563065	80.29550171	75.79370117	79.19676208	79.76883698	78.53195953	78.88621521	77.73145294	74.56821738
}\singletaskcrossqatable
\pgfplotstabletranspose[string type, colnames from=Language, input colnames to=Language]{\transposedsingletaskcrossqatable}{\singletaskcrossqatable}
\pgfplotstabletypeset[columns/Language/.style={string type, column name={Language / Locale ID}},columns/Validation/.style={numeric type,precision=1,zerofill},columns/Test/.style={numeric type,precision=1,zerofill},
  every head row/.style={
    before row={\toprule},
    after row={\midrule}
  },
  every last row/.style={before row={\midrule}, after row=\bottomrule}
]{\transposedsingletaskcrossqatable}
\caption{Zero-shot unseen languages: parameter-matched cross-lingual QA results.}
\label{tab:zeroshot-unseen-lang-param-matched-cqa}
\end{table}

\begin{table}
\pgfplotstableread{
Language as	bho	brx	gbm	gom	gu	hne	kn	mai	ml	mni	mr	mwr	or	pa	ps	sa	ur	Average
Validation 82.56031799	78.73670959	50.25761795	77.52978516	80.26475525	82.74514008	79.00293732	82.33808136	81.80185699	82.07558441	56.1841011	79.69996643	79.53029633	81.44894409	79.86673737	78.91400909	80.14323425	80.16475677	77.40360175
Test 82.51160431	78.65885925	46.96233368	75.35652924	78.61647797	82.60212708	79.83605957	82.45722961	79.91704559	80.96835327	58.97052002	80.74224854	77.89720154	81.08704376	81.71764374	80.5928421	81.64318848	79.94660187	77.24910609
}\singletaskcrossqatable
\pgfplotstabletranspose[string type, colnames from=Language, input colnames to=Language]{\transposedsingletaskcrossqatable}{\singletaskcrossqatable}
\pgfplotstabletypeset[columns/Language/.style={string type, column name={Language / Locale ID}},columns/Validation/.style={numeric type,precision=1,zerofill},columns/Test/.style={numeric type,precision=1,zerofill},
  every head row/.style={
    before row={\toprule},
    after row={\midrule}
  },
  every last row/.style={before row={\midrule}, after row=\bottomrule}
]{\transposedsingletaskcrossqatable}
\caption{Zero-shot unseen languages: \model cross-lingual QA results.}
\label{tab:zeroshot-unseen-lang-flora-cqa}
\end{table}

%% file: colm2024_conference.bbl
\begin{thebibliography}{28}
\providecommand{\natexlab}[1]{#1}
\providecommand{\url}[1]{\texttt{#1}}
\expandafter\ifx\csname urlstyle\endcsname\relax
  \providecommand{\doi}[1]{doi: #1}\else
  \providecommand{\doi}{doi: \begingroup \urlstyle{rm}\Url}\fi

\bibitem[Asai et~al.(2021)Asai, Kasai, Clark, Lee, Choi, and
  Hajishirzi]{asai-etal-2021-xor}
Akari Asai, Jungo Kasai, Jonathan Clark, Kenton Lee, Eunsol Choi, and Hannaneh
  Hajishirzi.
\newblock {XOR} {QA}: Cross-lingual open-retrieval question answering.
\newblock In Kristina Toutanova, Anna Rumshisky, Luke Zettlemoyer, Dilek
  Hakkani-Tur, Iz~Beltagy, Steven Bethard, Ryan Cotterell, Tanmoy Chakraborty,
  and Yichao Zhou (eds.), \emph{NAACL}, Online, June 2021. Association for
  Computational Linguistics.

\bibitem[Brown et~al.(2020)Brown, Mann, Ryder, Subbiah, Kaplan, Dhariwal,
  Neelakantan, Shyam, Sastry, Askell, Agarwal, Herbert-Voss, Krueger, Henighan,
  Child, Ramesh, Ziegler, Wu, Winter, Hesse, Chen, Sigler, Litwin, Gray, Chess,
  Clark, Berner, McCandlish, Radford, Sutskever, and
  Amodei]{NEURIPS2020_1457c0d6}
Tom Brown, Benjamin Mann, Nick Ryder, Melanie Subbiah, Jared~D Kaplan, Prafulla
  Dhariwal, Arvind Neelakantan, Pranav Shyam, Girish Sastry, Amanda Askell,
  Sandhini Agarwal, Ariel Herbert-Voss, Gretchen Krueger, Tom Henighan, Rewon
  Child, Aditya Ramesh, Daniel Ziegler, Jeffrey Wu, Clemens Winter, Chris
  Hesse, Mark Chen, Eric Sigler, Mateusz Litwin, Scott Gray, Benjamin Chess,
  Jack Clark, Christopher Berner, Sam McCandlish, Alec Radford, Ilya Sutskever,
  and Dario Amodei.
\newblock Language models are few-shot learners.
\newblock In H.~Larochelle, M.~Ranzato, R.~Hadsell, M.F. Balcan, and H.~Lin
  (eds.), \emph{Advances in Neural Information Processing Systems}, volume~33,
  pp.\  1877--1901. Curran Associates, Inc., 2020.
\newblock URL
  \url{https://proceedings.neurips.cc/paper_files/paper/2020/file/1457c0d6bfcb4967418bfb8ac142f64a-Paper.pdf}.

\bibitem[Caruana(1997)]{Caruana1997MultitaskL}
Rich Caruana.
\newblock Multitask learning.
\newblock \emph{Machine Learning}, 28:\penalty0 41--75, 1997.
\newblock URL \url{https://api.semanticscholar.org/CorpusID:45998148}.

\bibitem[Chronopoulou et~al.(2023)Chronopoulou, Pfeiffer, Maynez, Wang, Ruder,
  and Agrawal]{chronopoulou2023language}
Alexandra Chronopoulou, Jonas Pfeiffer, Joshua Maynez, Xinyi Wang, Sebastian
  Ruder, and Priyanka Agrawal.
\newblock Language and task arithmetic with parameter-efficient layers for
  zero-shot summarization, 2023.

\bibitem[Google et~al.(2023)Google, Anil, Dai, Firat, Johnson, Lepikhin,
  Passos, Shakeri, Taropa, Bailey, Chen, Chu, Clark, Shafey, Huang,
  Meier-Hellstern, Mishra, Moreira, Omernick, Robinson, Ruder, Tay, Xiao, Xu,
  Zhang, Abrego, Ahn, Austin, Barham, Botha, Bradbury, Brahma, Brooks, Catasta,
  Cheng, Cherry, Choquette-Choo, Chowdhery, Crepy, Dave, Dehghani, Dev, Devlin,
  Díaz, Du, Dyer, Feinberg, Feng, Fienber, Freitag, Garcia, Gehrmann,
  Gonzalez, Gur-Ari, Hand, Hashemi, Hou, Howland, Hu, Hui, Hurwitz, Isard,
  Ittycheriah, Jagielski, Jia, Kenealy, Krikun, Kudugunta, Lan, Lee, Lee, Li,
  Li, Li, Li, Li, Lim, Lin, Liu, Liu, Maggioni, Mahendru, Maynez, Misra,
  Moussalem, Nado, Nham, Ni, Nystrom, Parrish, Pellat, Polacek, Polozov, Pope,
  Qiao, Reif, Richter, Riley, Ros, Roy, Saeta, Samuel, Shelby, Slone, Smilkov,
  So, Sohn, Tokumine, Valter, Vasudevan, Vodrahalli, Wang, Wang, Wang, Wang,
  Wieting, Wu, Xu, Xu, Xue, Yin, Yu, Zhang, Zheng, Zheng, Zhou, Zhou, Petrov,
  and Wu]{anil2023palm}
Google, Rohan Anil, Andrew~M. Dai, Orhan Firat, Melvin Johnson, Dmitry
  Lepikhin, Alexandre Passos, Siamak Shakeri, Emanuel Taropa, Paige Bailey,
  Zhifeng Chen, Eric Chu, Jonathan~H. Clark, Laurent~El Shafey, Yanping Huang,
  Kathy Meier-Hellstern, Gaurav Mishra, Erica Moreira, Mark Omernick, Kevin
  Robinson, Sebastian Ruder, Yi~Tay, Kefan Xiao, Yuanzhong Xu, Yujing Zhang,
  Gustavo~Hernandez Abrego, Junwhan Ahn, Jacob Austin, Paul Barham, Jan Botha,
  James Bradbury, Siddhartha Brahma, Kevin Brooks, Michele Catasta, Yong Cheng,
  Colin Cherry, Christopher~A. Choquette-Choo, Aakanksha Chowdhery, Clément
  Crepy, Shachi Dave, Mostafa Dehghani, Sunipa Dev, Jacob Devlin, Mark Díaz,
  Nan Du, Ethan Dyer, Vlad Feinberg, Fangxiaoyu Feng, Vlad Fienber, Markus
  Freitag, Xavier Garcia, Sebastian Gehrmann, Lucas Gonzalez, Guy Gur-Ari,
  Steven Hand, Hadi Hashemi, Le~Hou, Joshua Howland, Andrea Hu, Jeffrey Hui,
  Jeremy Hurwitz, Michael Isard, Abe Ittycheriah, Matthew Jagielski, Wenhao
  Jia, Kathleen Kenealy, Maxim Krikun, Sneha Kudugunta, Chang Lan, Katherine
  Lee, Benjamin Lee, Eric Li, Music Li, Wei Li, YaGuang Li, Jian Li, Hyeontaek
  Lim, Hanzhao Lin, Zhongtao Liu, Frederick Liu, Marcello Maggioni, Aroma
  Mahendru, Joshua Maynez, Vedant Misra, Maysam Moussalem, Zachary Nado, John
  Nham, Eric Ni, Andrew Nystrom, Alicia Parrish, Marie Pellat, Martin Polacek,
  Alex Polozov, Reiner Pope, Siyuan Qiao, Emily Reif, Bryan Richter, Parker
  Riley, Alex~Castro Ros, Aurko Roy, Brennan Saeta, Rajkumar Samuel, Renee
  Shelby, Ambrose Slone, Daniel Smilkov, David~R. So, Daniel Sohn, Simon
  Tokumine, Dasha Valter, Vijay Vasudevan, Kiran Vodrahalli, Xuezhi Wang,
  Pidong Wang, Zirui Wang, Tao Wang, John Wieting, Yuhuai Wu, Kelvin Xu, Yunhan
  Xu, Linting Xue, Pengcheng Yin, Jiahui Yu, Qiao Zhang, Steven Zheng,
  Ce~Zheng, Weikang Zhou, Denny Zhou, Slav Petrov, and Yonghui Wu.
\newblock Palm 2 technical report, 2023.

\bibitem[Hu et~al.(2022)Hu, Shen, Wallis, Allen-Zhu, Li, Wang, Wang, and
  Chen]{hu2022lora}
Edward~J Hu, Yelong Shen, Phillip Wallis, Zeyuan Allen-Zhu, Yuanzhi Li, Shean
  Wang, Lu~Wang, and Weizhu Chen.
\newblock Lo{RA}: Low-rank adaptation of large language models.
\newblock In \emph{International Conference on Learning Representations}, 2022.
\newblock URL \url{https://openreview.net/forum?id=nZeVKeeFYf9}.

\bibitem[Huang et~al.(2023)Huang, Liu, Lin, Pang, Du, and
  Lin]{huang2023lorahub}
Chengsong Huang, Qian Liu, Bill~Yuchen Lin, Tianyu Pang, Chao Du, and Min Lin.
\newblock Lorahub: Efficient cross-task generalization via dynamic lora
  composition.
\newblock \emph{arXiv preprint arXiv:2307.13269}, 2023.

\bibitem[Ilharco et~al.(2022)Ilharco, Ribeiro, Wortsman, Gururangan, Schmidt,
  Hajishirzi, and Farhadi]{ilharco2022editing}
Gabriel Ilharco, Marco~Tulio Ribeiro, Mitchell Wortsman, Suchin Gururangan,
  Ludwig Schmidt, Hannaneh Hajishirzi, and Ali Farhadi.
\newblock Editing models with task arithmetic.
\newblock \emph{arXiv preprint arXiv:2212.04089}, 2022.

\bibitem[Jawahar et~al.(2023)Jawahar, Mukherjee, Liu, Kim, Abdul-Mageed,
  Lakshmanan, Awadallah, Bubeck, and Gao]{jawahar-etal-2023-automoe}
Ganesh Jawahar, Subhabrata Mukherjee, Xiaodong Liu, Young~Jin Kim, Muhammad
  Abdul-Mageed, Laks Lakshmanan, V.S., Ahmed~Hassan Awadallah, Sebastien
  Bubeck, and Jianfeng Gao.
\newblock {A}uto{M}o{E}: Heterogeneous mixture-of-experts with adaptive
  computation for efficient neural machine translation.
\newblock In Anna Rogers, Jordan Boyd-Graber, and Naoaki Okazaki (eds.),
  \emph{Findings of the Association for Computational Linguistics: ACL 2023},
  pp.\  9116--9132, Toronto, Canada, July 2023. Association for Computational
  Linguistics.
\newblock \doi{10.18653/v1/2023.findings-acl.580}.
\newblock URL \url{https://aclanthology.org/2023.findings-acl.580}.

\bibitem[Kudugunta et~al.(2021)Kudugunta, Huang, Bapna, Krikun, Lepikhin,
  Luong, and Firat]{kudugunta-etal-2021-beyond-distillation}
Sneha Kudugunta, Yanping Huang, Ankur Bapna, Maxim Krikun, Dmitry Lepikhin,
  Minh-Thang Luong, and Orhan Firat.
\newblock Beyond distillation: Task-level mixture-of-experts for efficient
  inference.
\newblock In Marie-Francine Moens, Xuanjing Huang, Lucia Specia, and Scott
  Wen-tau Yih (eds.), \emph{Findings of the Association for Computational
  Linguistics: EMNLP 2021}, pp.\  3577--3599, Punta Cana, Dominican Republic,
  November 2021. Association for Computational Linguistics.
\newblock \doi{10.18653/v1/2021.findings-emnlp.304}.
\newblock URL \url{https://aclanthology.org/2021.findings-emnlp.304}.

\bibitem[Lepikhin et~al.(2020)Lepikhin, Lee, Xu, Chen, Firat, Huang, Krikun,
  Shazeer, and Chen]{lepikhin2020gshard}
Dmitry Lepikhin, HyoukJoong Lee, Yuanzhong Xu, Dehao Chen, Orhan Firat, Yanping
  Huang, Maxim Krikun, Noam Shazeer, and Zhifeng Chen.
\newblock Gshard: Scaling giant models with conditional computation and
  automatic sharding, 2020.

\bibitem[Lester et~al.(2021)Lester, Al-Rfou, and
  Constant]{lester-etal-2021-power}
Brian Lester, Rami Al-Rfou, and Noah Constant.
\newblock The power of scale for parameter-efficient prompt tuning.
\newblock In Marie-Francine Moens, Xuanjing Huang, Lucia Specia, and Scott
  Wen-tau Yih (eds.), \emph{Proceedings of the 2021 Conference on Empirical
  Methods in Natural Language Processing}, Online and Punta Cana, Dominican
  Republic, November 2021. Association for Computational Linguistics.
\newblock URL \url{https://aclanthology.org/2021.emnlp-main.243}.

\bibitem[Lin et~al.(2021)Lin, Wu, Wang, and Li]{lin-etal-2021-learning}
Zehui Lin, Liwei Wu, Mingxuan Wang, and Lei Li.
\newblock Learning language specific sub-network for multilingual machine
  translation.
\newblock In Chengqing Zong, Fei Xia, Wenjie Li, and Roberto Navigli (eds.),
  \emph{ACL}, Online, August 2021. Association for Computational Linguistics.

\bibitem[Ouyang et~al.(2022)Ouyang, Wu, Jiang, Almeida, Wainwright, Mishkin,
  Zhang, Agarwal, Slama, Ray, Schulman, Hilton, Kelton, Miller, Simens, Askell,
  Welinder, Christiano, Leike, and Lowe]{Ouyang2022TrainingLM}
Long Ouyang, Jeff Wu, Xu~Jiang, Diogo Almeida, Carroll~L. Wainwright, Pamela
  Mishkin, Chong Zhang, Sandhini Agarwal, Katarina Slama, Alex Ray, John
  Schulman, Jacob Hilton, Fraser Kelton, Luke~E. Miller, Maddie Simens, Amanda
  Askell, Peter Welinder, Paul~Francis Christiano, Jan Leike, and Ryan~J. Lowe.
\newblock Training language models to follow instructions with human feedback.
\newblock \emph{ArXiv}, abs/2203.02155, 2022.
\newblock URL \url{https://api.semanticscholar.org/CorpusID:246426909}.

\bibitem[Pfeiffer et~al.(2020)Pfeiffer, Vuli{\'c}, Gurevych, and
  Ruder]{pfeiffer-etal-2020-mad}
Jonas Pfeiffer, Ivan Vuli{\'c}, Iryna Gurevych, and Sebastian Ruder.
\newblock {MAD-X}: {A}n {A}dapter-{B}ased {F}ramework for {M}ulti-{T}ask
  {C}ross-{L}ingual {T}ransfer.
\newblock In Bonnie Webber, Trevor Cohn, Yulan He, and Yang Liu (eds.),
  \emph{Proceedings of the 2020 Conference on Empirical Methods in Natural
  Language Processing (EMNLP)}, pp.\  7654--7673, Online, November 2020.
  Association for Computational Linguistics.
\newblock \doi{10.18653/v1/2020.emnlp-main.617}.
\newblock URL \url{https://aclanthology.org/2020.emnlp-main.617}.

\bibitem[Pfeiffer et~al.(2022)Pfeiffer, Goyal, Lin, Li, Cross, Riedel, and
  Artetxe]{pfeiffer-etal-2022-lifting}
Jonas Pfeiffer, Naman Goyal, Xi~Lin, Xian Li, James Cross, Sebastian Riedel,
  and Mikel Artetxe.
\newblock Lifting the curse of multilinguality by pre-training modular
  transformers.
\newblock In Marine Carpuat, Marie-Catherine de~Marneffe, and Ivan~Vladimir
  Meza~Ruiz (eds.), \emph{Proceedings of the 2022 Conference of the North
  American Chapter of the Association for Computational Linguistics: Human
  Language Technologies}, pp.\  3479--3495, Seattle, United States, July 2022.
  Association for Computational Linguistics.
\newblock \doi{10.18653/v1/2022.naacl-main.255}.
\newblock URL \url{https://aclanthology.org/2022.naacl-main.255}.

\bibitem[Pfeiffer et~al.(2023)Pfeiffer, Piccinno, Nicosia, Wang, Reid, and
  Ruder]{pfeiffer2023mmt5}
Jonas Pfeiffer, Francesco Piccinno, Massimo Nicosia, Xinyi Wang, Machel Reid,
  and Sebastian Ruder.
\newblock mmt5: Modular multilingual pre-training solves source language
  hallucinations, 2023.

\bibitem[Ruder et~al.(2023)Ruder, Clark, Gutkin, Kale, Ma, Nicosia, Rijhwani,
  Riley, Sarr, Wang, Wieting, Gupta, Katanova, Kirov, Dickinson, Roark,
  Samanta, Tao, Adelani, Axelrod, Caswell, Cherry, Garrette, Ingle, Johnson,
  Panteleev, and Talukdar]{ruder2023xtremeup}
Sebastian Ruder, Jonathan~H. Clark, Alexander Gutkin, Mihir Kale, Min Ma,
  Massimo Nicosia, Shruti Rijhwani, Parker Riley, Jean-Michel~A. Sarr, Xinyi
  Wang, John Wieting, Nitish Gupta, Anna Katanova, Christo Kirov, Dana~L.
  Dickinson, Brian Roark, Bidisha Samanta, Connie Tao, David~I. Adelani, Vera
  Axelrod, Isaac Caswell, Colin Cherry, Dan Garrette, Reeve Ingle, Melvin
  Johnson, Dmitry Panteleev, and Partha Talukdar.
\newblock Xtreme-up: A user-centric scarce-data benchmark for under-represented
  languages, 2023.

\bibitem[Shazeer et~al.(2017)Shazeer, Mirhoseini, Maziarz, Davis, Le, Hinton,
  and Dean]{shazeer2017outrageously}
Noam Shazeer, Azalia Mirhoseini, Krzysztof Maziarz, Andy Davis, Quoc Le,
  Geoffrey Hinton, and Jeff Dean.
\newblock Outrageously large neural networks: The sparsely-gated
  mixture-of-experts layer, 2017.

\bibitem[Soltan et~al.(2022)Soltan, Ananthakrishnan, FitzGerald, Gupta, Hamza,
  Khan, Peris, Rawls, Rosenbaum, Rumshisky, Prakash, Sridhar, Triefenbach,
  Verma, Tur, and Natarajan]{Soltan2022AlexaTM2F}
Saleh Soltan, Shankar Ananthakrishnan, Jack G.~M. FitzGerald, Rahul Gupta, Wael
  Hamza, Haidar Khan, Charith~S. Peris, Stephen Rawls, Andrew Rosenbaum, Anna
  Rumshisky, Chandan Prakash, Mukund Sridhar, Fabian Triefenbach, Apurv Verma,
  Gokhan Tur, and Premkumar Natarajan.
\newblock Alexatm 20b: Few-shot learning using a large-scale multilingual
  seq2seq model.
\newblock \emph{ArXiv}, abs/2208.01448, 2022.
\newblock URL \url{https://api.semanticscholar.org/CorpusID:251253416}.

\bibitem[Vu et~al.(2022)Vu, Barua, Lester, Cer, Iyyer, and
  Constant]{vu2022overcoming}
Tu~Vu, Aditya Barua, Brian Lester, Daniel Cer, Mohit Iyyer, and Noah Constant.
\newblock Overcoming catastrophic forgetting in zero-shot cross-lingual
  generation.
\newblock In \emph{EMNLP}, 2022.

\bibitem[Wang et~al.(2021)Wang, Tsvetkov, Ruder, and
  Neubig]{wang-etal-2021-efficient-test}
Xinyi Wang, Yulia Tsvetkov, Sebastian Ruder, and Graham Neubig.
\newblock Efficient test time adapter ensembling for low-resource language
  varieties.
\newblock In Marie-Francine Moens, Xuanjing Huang, Lucia Specia, and Scott
  Wen-tau Yih (eds.), \emph{EMNLP}, Punta Cana, Dominican Republic, November
  2021.

\bibitem[Wang et~al.(2023{\natexlab{a}})Wang, Wieting, and Clark]{wang2023fiat}
Xinyi Wang, John Wieting, and Jonathan~H. Clark.
\newblock Fiat: Fusing learning paradigms with instruction-accelerated tuning,
  2023{\natexlab{a}}.

\bibitem[Wang et~al.(2023{\natexlab{b}})Wang, Panda, Karlinsky, Feris, Sun, and
  Kim]{wang2023multitask}
Zhen Wang, Rameswar Panda, Leonid Karlinsky, Rogerio Feris, Huan Sun, and Yoon
  Kim.
\newblock Multitask prompt tuning enables parameter-efficient transfer
  learning.
\newblock In \emph{ICLR}, 2023{\natexlab{b}}.

\bibitem[Wang et~al.(2018)Wang, Dai, P{\'o}czos, and
  Carbonell]{Wang2018CharacterizingAA}
Zirui Wang, Zihang Dai, Barnab{\'a}s P{\'o}czos, and Jaime~G. Carbonell.
\newblock Characterizing and avoiding negative transfer.
\newblock \emph{2019 IEEE/CVF Conference on Computer Vision and Pattern
  Recognition (CVPR)}, pp.\  11285--11294, 2018.
\newblock URL \url{https://api.semanticscholar.org/CorpusID:53748459}.

\bibitem[Zadouri et~al.(2023)Zadouri, Üstün, Ahmadian, Ermiş, Locatelli, and
  Hooker]{zadouri2023loramoe}
Ted Zadouri, Ahmet Üstün, Arash Ahmadian, Beyza Ermiş, Acyr Locatelli, and
  Sara Hooker.
\newblock Pushing mixture of experts to the limit: Extremely parameter
  efficient moe for instruction tuning, 2023.

\bibitem[Zhang et~al.(2020)Zhang, Bapna, Sennrich, and
  Firat]{zhang-2020-share-mnmt}
Biao Zhang, Ankur Bapna, Rico Sennrich, and Orhan Firat.
\newblock Share or not? learning to schedule language-specific capacity for
  multilingual translation.
\newblock In \emph{ICLR}, 2020.

\bibitem[Zhu et~al.(2023)Zhu, Wichers, Lin, Wang, Chen, Shu, Lu, Liu, Luo,
  Chen, et~al.]{zhu2023sira}
Yun Zhu, Nevan Wichers, Chu-Cheng Lin, Xinyi Wang, Tianlong Chen, Lei Shu, Han
  Lu, Canoee Liu, Liangchen Luo, Jindong Chen, et~al.
\newblock Sira: Sparse mixture of low rank adaptation.
\newblock \emph{arXiv preprint arXiv:2311.09179}, 2023.

\end{thebibliography}
